\pdfoutput=1

\documentclass[journal]{IEEEtran}

\usepackage[noadjust]{cite}
\usepackage[pdftex]{graphicx}
\graphicspath{{Figures/}}
\usepackage{multirow}
\usepackage{xcolor}
\usepackage{pifont}
\usepackage{array}
\usepackage{amssymb}
\usepackage{url}

\begin{document}

\title{Biometric Template Protection for Neural-Network-based Face Recognition Systems:\\A Survey of Methods and Evaluation Techniques}

\author{Vedrana~Krivoku\'ca~Hahn
        and~S\'ebastien~Marcel \\
        \textit{Idiap Research Institute (Martigny, Switzerland)}  
}

\maketitle

\begin{abstract}

As automated face recognition applications tend towards ubiquity, there is a growing need to secure the sensitive face data used within these systems. This paper presents a survey of biometric template protection (BTP) methods proposed for securing face ``templates'' (images/features) in neural-network-based face recognition systems. The BTP methods are categorised into two types: Non-NN and NN-learned. Non-NN methods use a neural network (NN) as a feature extractor, but the BTP part is based on a non-NN algorithm applied at either image-level or feature-level. In contrast, NN-learned methods specifically employ a NN to learn a protected  template from the unprotected face image/features. We present examples of Non-NN and NN-learned face BTP methods from the literature, along with a discussion of the two categories' comparative strengths and weaknesses. We also investigate the techniques used to evaluate these BTP methods, in terms of the three most common BTP criteria: ``recognition accuracy'', ``irreversibility'', and ``renewability/unlinkability''. As expected, the recognition accuracy of protected face recognition systems is generally evaluated using the same (empirical) techniques employed for evaluating standard (unprotected) biometric systems. On the contrary, most irreversibility and renewability/unlinkability evaluations are found to be based on theoretical assumptions/estimates or verbal implications, with a lack of empirical validation in a practical face recognition context. We recommend, therefore, a greater focus on empirical evaluation strategies, to provide more concrete insights into the irreversibility and renewability/unlinkability of face BTP methods in practice. Additionally, an exploration of the reproducibility of the studied BTP works, in terms of the public availability of their implementation code and evaluation datasets/procedures, suggests that it would currently be difficult for the BTP community to faithfully replicate (and thus validate) most of the reported findings. So, we advocate for a push towards reproducibility, in the hope of furthering our understanding of the face BTP research field.

\end{abstract}


\section{Introduction}

\IEEEPARstart{A}utomated face recognition is fast becoming a default solution for identity management in our increasingly interconnected society. Applications range from single-person access control (e.g., phone/computer login) to large-scale operations that process thousands of people on a daily basis (e.g., passport control at an airport). The widespread adoption of this technology is due to its ability to simultaneously provide a high level of security and convenience to its users. Nevertheless, it is important to consider how our privacy may be impacted by the collection and use of our sensitive face data by these numerous face recognition applications. 

Modern face recognition systems are commonly based on deep learning architectures, which train a multi-layer neural network to learn a set of representative face features from images of a person's face. Recently, it has been shown that these features can be ``inverted'' to recover an approximation of the original face image \cite{zs16, c17, mc18, os22}, and that they can reveal additional information about the underlying person (e.g., gender, race, age) \cite{f20, tf20a, tf20}. These findings indicate that the face features used by deep-learning-based face recognition systems are rich with personally identifiable information, which may represent a threat to the privacy of the systems' users if their face features are leaked to an adversary. This is especially problematic considering frameworks like the EU General Data Protection Regulation (GDPR)\footnote{\url{https://bit.ly/2VGSa5e}}, which imposes a legal obligation to handle biometric data carefully to protect individuals' digital identities. So, the issue of how to protect our face data in automated face recognition systems is becoming pressing in response to these systems' increasingly widespread adoption. 

Fortunately, the past decade has witnessed an enrichment of the Biometric Template Protection (BTP) research field, which strives to develop effective ways of protecting our biometric ``templates'' (images or features). This paper conducts a survey of \textit{face} template protection methods for \textit{neural-network-based} face recognition systems. In particular, we present a categorisation of the methods that have been proposed for face template protection thus far: \textit{Non-NN} methods use a neural network (NN) as a feature extractor, but the BTP part is \textit{not} defined by a NN (i.e., the BTP algorithm is ``handcrafted'' or based on non-NN machine-learning), while \textit{NN-learned} methods specifically employ a NN to \textit{learn} a protected template from the unprotected face template. We also present a thorough investigation into the techniques used to evaluate these methods' abilities to satisfy the three most important criteria of BTP schemes\footnote{\url{https://bit.ly/398Az9k}}: ``recognition accuracy'', ``irreversibility'', and ``renewability/unlinkability''. Additionally, we shed light on the reproducibility of the investigated works by exploring the public availability of the method implementations and the datasets/procedures used to evaluate them. It is our hope that this survey will help to clarify the state of the face BTP research field, thereby motivating further advancements towards face recognition systems that are better able to protect the precious face templates of their users. 

To help the reader find their bearings, Table I summarises the studied face BTP works, which are categorised into \textit{Non-NN} and \textit{NN-learned} approaches, then further classified into sub-categories describing the high-level type of BTP method (details provided in Section II). Table I also presents an overview of which BTP criteria were evaluated and how, for each of the investigated BTP works (details provided in Section III).

\begin{table*}[!ht]
\renewcommand{\arraystretch}{1.2}
\caption{Summary of the studied face BTP approaches and their corresponding evaluation techniques.} 
\centering
\begin{tabular}{|c|c|c|c|c|c|c|c|c|}
\hline
\multirow{3}{*}{\textbf{Category}} & \multirow{3}{*}{\textbf{Sub-category}} & \multirow{3}{*}{\textbf{Reference}} & \multicolumn{6}{c|}{\textbf{Evaluation of BTP criteria:}} \\
\cline{4-9}
 & & & \textbf{Recognition} &  \multicolumn{2}{c|}{\textbf{Irreversibility}} & \multicolumn{3}{c|}{\textbf{Renewability/Unlinkability}} \\
\cline{5-9}
 & & & \textbf{accuracy} & \textbf{Theoretical} & \textbf{Empirical} & \textbf{Implication} & \textbf{Theoretical} & \textbf{Empirical} \\
 \hline

\multirow{26}{*}{Non-NN} & Homomorphic encryption & \cite{m17} & {\color{olive}\ding{52}} & {\color{olive}\ding{52}} &  {\color{red}\ding{56}} & {\color{red}\ding{56}} & {\color{red}\ding{56}} & {\color{red}\ding{56}} \\

 \cline{4-9}
 
 & Homomorphic encryption & \cite{b18} & {\color{olive}\ding{52}} & {\color{olive}\ding{52}} & {\color{red}\ding{56}} & {\color{olive}\ding{52}} & {\color{red}\ding{56}} & {\color{red}\ding{56}} \\
 
 \cline{4-9}
 
 & Homomorphic encryption & \cite{b19} & {\color{olive}\ding{52}} & {\color{olive}\ding{52}} & {\color{red}\ding{56}} & {\color{olive}\ding{52}} & {\color{red}\ding{56}} & {\color{red}\ding{56}} \\
 
 \cline{4-9}
 
 & Homomorphic encryption & \cite{j20} & {\color{olive}\ding{52}} & {\color{olive}\ding{52}} & {\color{red}\ding{56}} & {\color{olive}\ding{52}} & {\color{red}\ding{56}} & {\color{red}\ding{56}} \\
 
 \cline{4-9}
 
 & Homomorphic encryption & \cite{d21a, d21} & {\color{olive}\ding{52}} & {\color{olive}\ding{52}} & {\color{red}\ding{56}} & {\color{olive}\ding{52}} & {\color{red}\ding{56}} & {\color{red}\ding{56}} \\

 \cline{4-9}
 
 & Homomorphic encryption & \cite{e20, e22} & {\color{olive}\ding{52}} & {\color{olive}\ding{52}} & {\color{red}\ding{56}} & {\color{olive}\ding{52}} & {\color{red}\ding{56}} & {\color{red}\ding{56}} \\

 \cline{4-9}
 
 & Homomorphic encryption & \cite{o21} & {\color{olive}\ding{52}} & {\color{olive}\ding{52}} & {\color{red}\ding{56}} & {\color{olive}\ding{52}} & {\color{red}\ding{56}} & {\color{red}\ding{56}} \\

 \cline{4-9}
 
 & Hashing & \cite{p17} & {\color{olive}\ding{52}} & {\color{olive}\ding{52}} & {\color{red}\ding{56}} & {\color{red}\ding{56}} & {\color{red}\ding{56}} & {\color{red}\ding{56}} \\
 
 \cline{4-9}
 
 & Hashing & \cite{m19} & {\color{olive}\ding{52}} & {\color{red}\ding{56}} & {\color{red}\ding{56}} & {\color{olive}\ding{52}} & {\color{red}\ding{56}} & {\color{red}\ding{56}} \\
 
 \cline{4-9}
 
 & Hashing & \cite{ko20, ko21} & {\color{red}\ding{56}} & {\color{red}\ding{56}} & {\color{olive}\ding{52}} & {\color{red}\ding{56}} & {\color{red}\ding{56}} & {\color{olive}\ding{52}} \\
 
 \cline{4-9}
 
 & Hashing & \cite{g19} & {\color{olive}\ding{52}} & {\color{olive}\ding{52}} & {\color{red}\ding{56}} & {\color{red}\ding{56}} & {\color{olive}\ding{52}} & {\color{red}\ding{56}} \\
 
 \cline{4-9}
 
 & Hashing & \cite{r21, r22} & {\color{olive}\ding{52}} & {\color{red}\ding{56}} & {\color{olive}\ding{52}} & {\color{red}\ding{56}} & {\color{olive}\ding{52}} & {\color{red}\ding{56}} \\
 
 \cline{4-9}
 
 & Hashing & \cite{a14} & {\color{olive}\ding{52}} & {\color{red}\ding{56}} & {\color{red}\ding{56}} & {\color{red}\ding{56}} & {\color{red}\ding{56}} & {\color{red}\ding{56}} \\
 
 \cline{4-9}
 
 & Feature transformation & \cite{a19} & {\color{olive}\ding{52}} & {\color{red}\ding{56}} & {\color{olive}\ding{52}} & {\color{red}\ding{56}} & {\color{red}\ding{56}} & {\color{red}\ding{56}} \\
 
 \cline{4-9}
 
 & Hashing & \cite{dm20} & {\color{olive}\ding{52}} & {\color{red}\ding{56}} & {\color{red}\ding{56}} & {\color{red}\ding{56}} & {\color{red}\ding{56}} & {\color{red}\ding{56}} \\
 
 \cline{4-9}
 
 & Image distortion & \cite{s18} & {\color{olive}\ding{52}} & {\color{olive}\ding{52}} & {\color{red}\ding{56}} & {\color{red}\ding{56}} & {\color{red}\ding{56}} & {\color{red}\ding{56}} \\
 
 \cline{4-9}
 
 & Image distortion & \cite{k20} & {\color{olive}\ding{52}} & {\color{red}\ding{56}} & {\color{olive}\ding{52}} & {\color{red}\ding{56}} & {\color{red}\ding{56}} & {\color{olive}\ding{52}} \\
 
 \cline{4-9}
 
 & Hashing & \cite{d19} & {\color{olive}\ding{52}} & {\color{olive}\ding{52}} & {\color{red}\ding{56}} & {\color{red}\ding{56}} & {\color{red}\ding{56}} & {\color{olive}\ding{52}} \\
 
 \cline{4-9}
 
 & Hashing & \cite{d20} & {\color{olive}\ding{52}} & {\color{olive}\ding{52}} & {\color{red}\ding{56}} & {\color{olive}\ding{52}} & {\color{red}\ding{56}} & {\color{red}\ding{56}} \\
 
 \cline{4-9}
 
 & Hashing & \cite{w20} & {\color{olive}\ding{52}} & {\color{olive}\ding{52}} & {\color{red}\ding{56}} & {\color{olive}\ding{52}} & {\color{red}\ding{56}} & {\color{red}\ding{56}} \\
 
 \cline{4-9}
 
 & Hashing & \cite{a20} & {\color{olive}\ding{52}} & {\color{red}\ding{56}} & {\color{red}\ding{56}} & {\color{red}\ding{56}} & {\color{red}\ding{56}} & {\color{red}\ding{56}} \\
 
 \cline{4-9}
 
 & Hashing & \cite{x20} & {\color{olive}\ding{52}} & {\color{red}\ding{56}} & {\color{red}\ding{56}} & {\color{red}\ding{56}} & {\color{red}\ding{56}} & {\color{olive}\ding{52}} \\
 
 \cline{4-9}
 
 & Hashing & \cite{j21} & {\color{olive}\ding{52}} & {\color{olive}\ding{52}} & {\color{red}\ding{56}} & {\color{red}\ding{56}} & {\color{olive}\ding{52}} & {\color{red}\ding{56}} \\
 
 \cline{4-9}
 
 & Hashing & \cite{h21} & {\color{olive}\ding{52}} & {\color{olive}\ding{52}} & {\color{red}\ding{56}} & {\color{olive}\ding{52}} & {\color{red}\ding{56}} & {\color{red}\ding{56}} \\
 
 \cline{4-9}
 
 & Feature transformation & \cite{p19} & {\color{olive}\ding{52}} & {\color{olive}\ding{52}} & {\color{red}\ding{56}} & {\color{olive}\ding{52}} & {\color{red}\ding{56}} & {\color{red}\ding{56}} \\
 
 \cline{4-9}
 
 & Feature transformation & \cite{kh21, kh22} & {\color{olive}\ding{52}} & {\color{olive}\ding{52}} & {\color{olive}\ding{52}} & {\color{red}\ding{56}} & {\color{red}\ding{56}} & {\color{olive}\ding{52}} \\
 
\hline

\multirow{14}{*}{NN-learned} & Learns pre-defined template & \cite{p15, p16} & {\color{olive}\ding{52}} & {\color{olive}\ding{52}} & {\color{olive}\ding{52}} & {\color{olive}\ding{52}} & {\color{red}\ding{56}} & {\color{red}\ding{56}} \\

 \cline{4-9}
 
 & Learns pre-defined template & \cite{z19} & {\color{olive}\ding{52}} & {\color{olive}\ding{52}} & {\color{olive}\ding{52}} & {\color{olive}\ding{52}} & {\color{red}\ding{56}} & {\color{red}\ding{56}} \\
 
 \cline{4-9}
 
 & Learns pre-defined template & \cite{j18} & {\color{olive}\ding{52}} & {\color{olive}\ding{52}} & {\color{olive}\ding{52}} & {\color{olive}\ding{52}} & {\color{red}\ding{56}} & {\color{red}\ding{56}} \\
 
 \cline{4-9}
 
 & Learns pre-defined template & \cite{jc19} & {\color{olive}\ding{52}} & {\color{olive}\ding{52}} & {\color{olive}\ding{52}} & {\color{olive}\ding{52}} & {\color{red}\ding{56}} & {\color{red}\ding{56}} \\
 
 \cline{4-9}
 
 & Learns pre-defined template & \cite{j19} & {\color{olive}\ding{52}} & {\color{olive}\ding{52}} & {\color{red}\ding{56}} & {\color{olive}\ding{52}} & {\color{red}\ding{56}} & {\color{red}\ding{56}} \\
 
 \cline{4-9}
 
 & Learns own template & \cite{r18} & {\color{olive}\ding{52}} & {\color{red}\ding{56}} & {\color{red}\ding{56}} & {\color{red}\ding{56}} & {\color{red}\ding{56}} & {\color{red}\ding{56}} \\
 
 \cline{4-9}
 
 & Learns own template & \cite{z20} & {\color{olive}\ding{52}} & {\color{olive}\ding{52}} & {\color{red}\ding{56}} & {\color{red}\ding{56}} & {\color{red}\ding{56}} & {\color{red}\ding{56}} \\
 
 \cline{4-9}
 
 & Learns own template & \cite{t19} & {\color{olive}\ding{52}} & {\color{olive}\ding{52}} & {\color{olive}\ding{52}} & {\color{red}\ding{56}} & {\color{red}\ding{56}} & {\color{red}\ding{56}} \\
 
 \cline{4-9}
 
 & Learns own template & \cite{s21} & {\color{olive}\ding{52}} & {\color{red}\ding{56}} & {\color{olive}\ding{52}} & {\color{olive}\ding{52}} & {\color{red}\ding{56}} & {\color{red}\ding{56}} \\
 
 \cline{4-9}
 
 & Learns own template & \cite{c19} & {\color{olive}\ding{52}} & {\color{olive}\ding{52}} & {\color{red}\ding{56}} & {\color{olive}\ding{52}} & {\color{red}\ding{56}} & {\color{red}\ding{56}} \\
 
 \cline{4-9}
 
 & Learns own template & \cite{m21} & {\color{olive}\ding{52}} & {\color{olive}\ding{52}} & {\color{red}\ding{56}} & {\color{red}\ding{56}} & {\color{red}\ding{56}} & {\color{olive}\ding{52}} \\
 
 \cline{4-9}
 
 & Learns own template & \cite{p21} & {\color{olive}\ding{52}} & {\color{olive}\ding{52}} & {\color{red}\ding{56}} & {\color{red}\ding{56}} & {\color{red}\ding{56}} & {\color{olive}\ding{52}} \\
 
 \cline{4-9}
 
 & Learns own template & \cite{c20} & {\color{olive}\ding{52}} & {\color{olive}\ding{52}} & {\color{red}\ding{56}} & {\color{olive}\ding{52}} & {\color{red}\ding{56}} & {\color{red}\ding{56}} \\
 
 \cline{4-9}
 
 & Learns own template & \cite{l21} & {\color{olive}\ding{52}} & {\color{olive}\ding{52}} & {\color{red}\ding{56}} & {\color{red}\ding{56}} & {\color{olive}\ding{52}} & {\color{olive}\ding{52}} \\
 
\hline
\end{tabular}
\end{table*} 

To the best of our knowledge, no survey paper of this nature currently exists in the literature. Although a number of BTP surveys have emerged over the past couple of decades (e.g., \cite{r11, sp17, mnk20}), their focus and contributions differ significantly from ours. For example, whereas our work focuses specifically (and deliberately) on \textit{face} template protection, other surveys tend to be more general, preferring to spread their focus across multiple biometric modalities (e.g., fingerprints, iris, voice, face, finger veins, palm prints, etc.) instead of honing in on a single modality. Furthermore, our paper proposes the \textit{first explicit categorisation} of the existing face BTP methods into those that are \textit{not} learned by a neural network (``Non-NN'') and those that \textit{are} (``NN-learned''), with an in-depth discussion of both categories. On the contrary, most existing surveys focus exclusively on non-NN methods (a single NN-learned method is mentioned briefly in \cite{mnk20}). Although we did come across a paper that concentrates solely on the face modality and cites works belonging to both types of BTP categories \cite{dc20}, the paper does \textit{not} focus specifically on surveying the BTP techniques themselves but rather on providing an overview of the processes used to generate cryptographic keys from faces -- in fact, the studied key generation methods are categorised according to the type of \textit{feature extractor} used as opposed to the adopted BTP method (if any), and only ten papers in total are considered in this work. So, as far as we are aware, our survey is unique in its focused and comprehensive coverage of both non-NN and NN-learned face BTP methods. Finally, to the best of our knowledge, ours is the only survey paper to present a dedicated investigation into the techniques used to evaluate the studied BTP methods, for \textit{all three} BTP criteria (recognition accuracy, irreversibility, and renewability/unlinkability). We believe, therefore, that our work serves as an effective up-to-date reference on the state of the face BTP research field.  This may be considered an important contribution to both the BTP research community and any practitioners looking for a reasonable starting point when considering ways of protecting face templates in real-world face recognition systems.

It should be noted that the ``biometric template protection'' (BTP) research field (which is the focus of our survey paper) is different to the field of ``privacy-enhancing biometrics'', for which an excellent survey can be found in \cite{mr21}\footnote{In this paper, the authors explicitly state that the topic of biometric template protection is not covered by their definition of biometric privacy-enhancing techniques, as it follows different assumptions.}.  Although both topics fundamentally aim to preserve the privacy of the biometric system users, the approaches for doing so are different.  The main focus of BTP methods is on transforming biometric templates to secure the templates themselves, such that the original template cannot be recovered from its protected counterpart.  If this can be achieved, then both the privacy of the template owners (i.e., the users enrolled in the biometric recognition system) and the security of the system itself, can be ensured.  So, in the BTP field, we consider the biometric template as an \textit{entire} data construct that must be protected.  Biometric privacy-enhancing techniques, on the other hand, target the protection of \textit{specific biometric attributes}, such as gender, age, and ethnicity.  In this case, the aim is usually to remove these attributes from the biometric data or to obfuscate them in some way, thereby limiting the amount of privacy-sensitive information that can be extracted and used for recognition purposes.  Having made this comparison between the two research directions, we further note that the fields of BTP and privacy-enhancing biometrics are complementary, so our proposed survey on BTP methods in neural-network based face recognition systems may be considered complementary to the survey on privacy-preserving face biometrics in \cite{mr21}\footnote{However, unlike our survey, \cite{mr21} does not focus explicitly on \textit{neural-network-based} face recognition systems.}.  

The remainder of this paper is structured as follows. Section II proposes a classification of the existing face BTP methods into two types (\textit{Non-NN} and \textit{NN-learned}), presents examples from the literature of each type of method, and discusses the strengths and weaknesses of the two categories of approaches. Section III investigates the techniques used to evaluate the methods from Section II with respect to the three most important BTP criteria (recognition accuracy, irreversibility, and renewability/unlinkability), beginning with a consideration of the reproducibility of these works in terms of their public availability. Finally, Section IV concludes this survey with a summary of our main findings.

\section{Face BTP Methods}
  
This section considers the types of methods proposed to protect face data in automated face recognition systems. \textit{Face data} refers to either the face \textit{image} or a set of representative \textit{features} extracted from the image, depending on whether the protection mechanism is applied at image-level or feature-level. To comply with the common terminology, we shall henceforth refer to these protection methods as Biometric Template Protection (BTP) methods, where the word \textit{template} is synonymous with the aforementioned \textit{face data}.  

In the context of this survey, we are only interested in BTP methods proposed for face recognition systems that incorporate a \textit{Neural Network} (NN) (or, more commonly, a \textit{Deep Neural Network} (DNN)) into their pipeline.  Based on our perusal of the relevant literature, we propose a categorisation of the face BTP approaches into two main types, according to what role the NN plays in the face recognition system's pipeline: ``Non-NN'' and ``NN-learned''.  

\textit{Non-NN} BTP methods do \textit{not} use a neural network to define the BTP algorithm; rather, the NN serves simply as a \textit{feature extractor}.  The NN can be used as either a \textit{pre}-BTP or \textit{post}-BTP feature extractor. In the \textit{pre}-BTP case, the NN would first be used to extract a set of face features (i.e., a \textit{feature vector}) from the face image, then the BTP algorithm would be applied to this feature vector (i.e., at \textit{feature-level}) to generate the protected face template. In the \textit{post}-BTP case, the BTP algorithm would be applied directly to the face image (i.e., at \textit{image-level}), following which the NN would be used to extract features from the \textit{protected} image. Regardless of the NN's precise positioning as a feature extractor, the important point is that the NN does not form part of the BTP method itself; instead, Non-NN BTP methods are defined in terms of either handcrafted or non-NN machine-learning algorithms. 

\textit{NN-learned} BTP methods, on the other hand, \textit{do} use a neural network to express the BTP algorithm. More specifically, a NN (usually a DNN) is used to \textit{learn} the protected template either directly from its corresponding face image or from the extracted feature vector. In the former case, where the BTP is applied at \textit{image-level}, the NN would be trained in an end-to-end manner (i.e., image $\rightarrow$ protected template). In the latter case, where the BTP is applied at \textit{feature-level}, the training would start from the extracted face features (i.e., feature vector $\rightarrow$ protected face template). The feature extraction process could be conducted either using a separate NN (as for Non-NN BTP methods) or a handcrafted/machine-learned (non-NN) feature extraction algorithm. Regardless of how the face features are extracted or the stage at which the BTP method is applied, the important point is that NN-learned BTP methods are not explicitly formulated; instead, we rely on a NN to \textit{learn} the BTP algorithm that will transform a face template (image or feature vector) to its protected template.

Fig. 1 illustrates the Non-NN and NN-learned BTP method types, sub-divided into ``feature-level'' and ``image-level'' depending on the stage at which the BTP method is applied. Note that all outputs in Fig. 1 are the ``protected face template''; however, when the protected template comes from a NN, the output could alternatively be a \textit{classification} of the template into its corresponding identity, instead of the template itself. As this may be considered an implementation detail (depending on what the NN was trained to output), and since the main aim of any BTP method is to generate a protected template from the unprotected template, Fig. 1 has been simplified to always output the ``protected face template''. 

\begin{figure}[!ht]
\centering
\includegraphics[width=0.8\columnwidth]{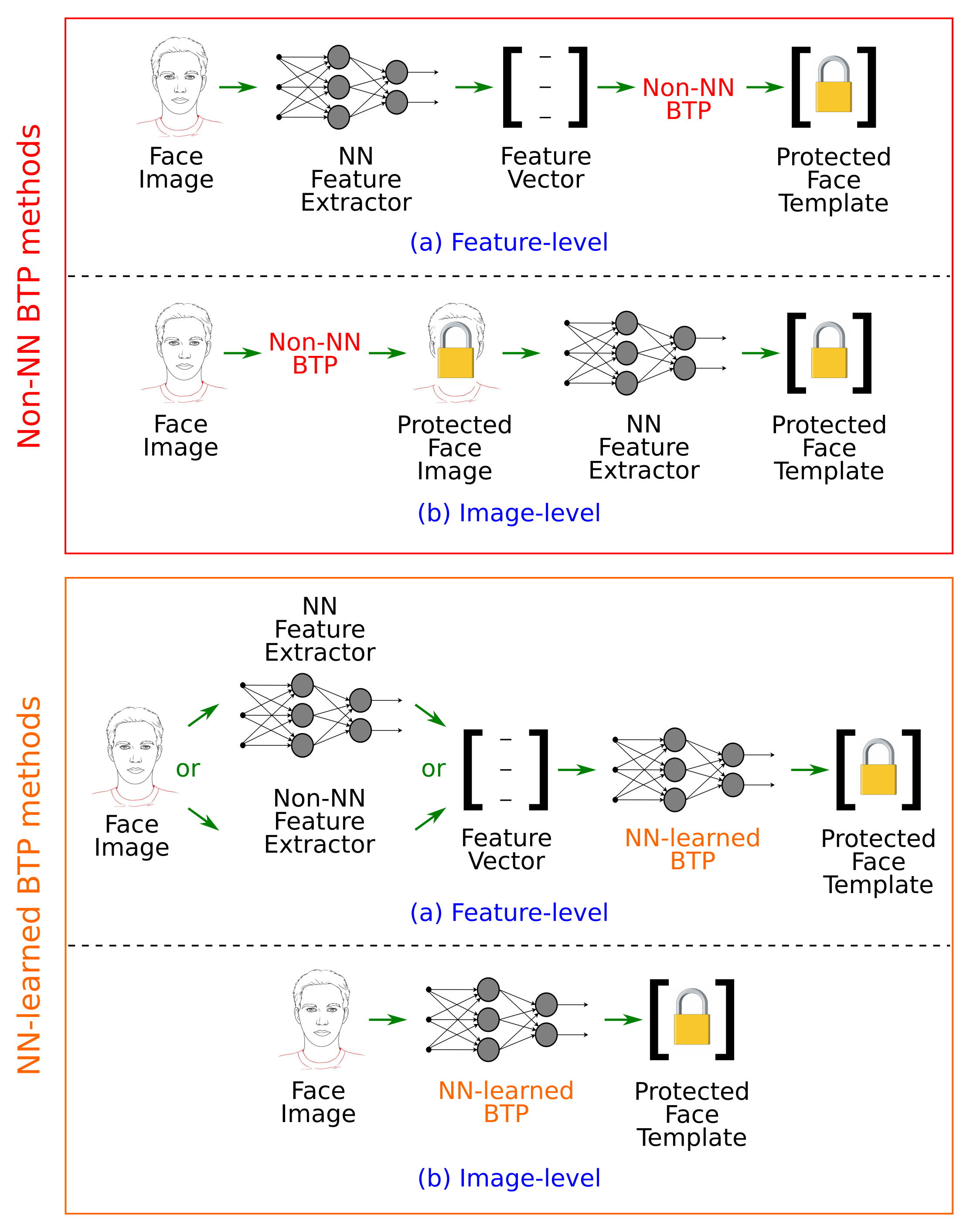}%
\caption{The Non-NN and NN-learned face BTP method types. Each type of BTP method can be applied at either feature-level or image-level.}
\end{figure}

Sections II.A and II.B present examples from the literature of Non-NN and NN-learned face BTP methods, respectively.

\subsection{Non-NN Face BTP Methods}

Examples of Non-NN face BTP methods in the literature, include: \cite{m17, b18, b19, j20, d21a, d21, e20, e22, o21, p17, m19, ko20, ko21, g19, r21, r22, a14, a19, dm20, s18, k20, d19, d20, w20, a20, x20, j21, h21, p19, kh21, kh22}. The overwhelming majority of these works (\cite{m17, b18, b19, j20, d21a, d21, e20, e22, o21, p17, m19, ko20, ko21, g19, r21, r22, a14, a19, dm20, d19, d20, w20, a20, x20, j21, h21, p19, kh21, kh22}) focus on applying a BTP method at \textit{feature-level} (as per Fig. 1), while only a couple of publications (\cite{s18, k20}) consider \textit{image-level} BTP methods. This preference for feature-level BTP may be attributed to the availability of several pre-trained face recognition DNN models, which have been shown to be capable of extracting highly discriminative face features from face images. Consequently, the task of applying a protection mechanism on top of robust face features may be more appealing than the potentially more challenging task of designing a good image-level BTP method \textit{plus} a robust feature extractor for the protected images -- since features extracted from protected images would likely lie in a different domain than features extracted from unprotected images, it is not clear whether existing feature extraction algorithms would be applicable to protected face images. 

Of the Non-NN BTP methods applied at feature-level, one of the most commonly studied protection algorithms is \textit{homomorphic encryption} (HE), which was proposed in \cite{m17, b18, b19, j20, d21a, d21, e20, e22, o21} for the protection of face features that have been extracted from face images using pre-trained DNN models. Unlike traditional encryption techniques, HE allows us to perform operations on \textit{encrypted} data without having to first decrypt it. In the context of face BTP, HE would allow us to compare reference and probe encrypted templates (generated during the enrollment and authentication stages, respectively) directly in the protected domain. This is in contrast to traditional encryption methods, which would necessitate comparison in the decrypted (i.e., unprotected) domain. Another important advantage of using HE for BTP is that, theoretically, there should be no loss in the resulting recognition accuracy, since the comparison score obtained in the encrypted domain should be the same as that obtained in the original (unprotected) domain. Although these important benefits of HE have long been known, its computational complexity has rendered it impractical for application in face BTP until fairly recently. Current work in this area tends to focus on finding a balance between speeding up HE operations (e.g., by quantising/binarising the face templates or reducing their dimensionality) while simultaneously minimising losses in the resulting recognition accuracy (e.g., by trying to encrypt face templates in their original -- usually floating-point -- domain). Although \cite{m17, b18, b19, j20, d21a, d21, e20, e22, o21} demonstrate important advances towards these goals for HE applied to both face \textit{verification} (1-to-1 comparison) \cite{m17, b18, j20} and \textit{identification} (1-to-\textit{N} comparison) \cite{b19, d21a, d21, e20, e22, o21}, the greatest disadvantage of HE is that the encrypted templates remain secure only insofar as the corresponding decryption key remains secret; if an adversary gains access to this secret key, they can completely reverse the protection algorithm to obtain the original (unencrypted) template. 

The invertible nature of encryption algorithms makes them less attractive for securing sensitive data like face templates, compared to non-invertible operations like one-way \textit{cryptographic hash functions}. A cryptographic hash function creates a fixed-size, predictable output called a ``hash'' from each instance of the same input data (e.g., a password), such that it is mathematically impossible to recover the original input from its computed hash. Despite the appeal of this ``one-wayness'' property, it is difficult to apply cryptographic hash functions to face templates due to the intra-class variability that is inherent in biometric measurements. This is because hash functions are purposely designed to exaggerate small differences in the input, such that even a tiny change would result in a completely different output hash; however, different instances of the same person's face template, such as would be acquired during enrollment and each authentication attempt, will never be identical, so comparing reference and probe face templates in the hashed domain would be practically impossible. 

Considering the sensitivity of cryptographic hash functions to small changes in the input, face BTP methods that employ cryptographic hashing often avoid creating hashes from the face templates themselves. Instead, they apply the hashing algorithm to randomly generated, external codewords, which are related to the face templates by some sort of mathematical function or mapping. A well-known example of such a method is the Fuzzy Commitment scheme \cite{j99}, whereby a random, user-specific codeword is ``bound'' with the user's biometric template by calculating the difference between the two entities (i.e., $\texttt{binding} = \texttt{template} - \texttt{codeword}$). The binding calculated from the reference template and chosen codeword during enrollment constitutes the \textit{protected} reference template (i.e., $\texttt{protected reference template} = \texttt{reference template} - \texttt{codeword}$), which is stored in the template database along with a cryptographic hash of the codeword. During authentication, the aim is to ``unbind'' the protected reference template to recover this codeword (i.e., $\texttt{recovered codeword} = \texttt{probe template} - \texttt{protected reference template}$). This should be possible provided that the probe template is \textit{close enough} to the reference template (to which the codeword is bound). In this case, an appropriate error-correction mechanism should be able to correct a sufficient number of errors in the recovered codeword, such that the hash of the corrected codeword matches the hash of the reference codeword. Variants of the Fuzzy Commitment scheme were investigated for the protection of NN-extracted face features in \cite{m19, g19, ko20, ko21}. 

Another well-known ``binding'' type of BTP method is the Fuzzy Vault scheme \cite{j06}, where the randomly chosen, user-specific codeword serves as a set of coefficients defining a secret polynomial, and the elements of the user's biometric template are the polynomial's inputs. The outputs resulting from the polynomial mapping (i.e., ``true points''), plus a set of ``chaff points'' added to hide the true points, constitute the protected template (i.e., \textit{fuzzy vault}), along with a cryptographic hash of the random codeword (i.e., polynomial coefficients). During authentication, the probe biometric template is used to determine the true points and thereby recover the secret polynomial from the reference fuzzy vault, which will be successful if the cryptographic hash of the recovered set of polynomial coefficients matches the hash of the set of reference coefficients. The Fuzzy Vault scheme was proposed as a Non-NN face BTP method in \cite{r21, r22}. 

Other examples of Non-NN face BTP methods that cryptographically hash a random, user-specific codeword instead of the face template itself, include \cite{d20, j21}. In \cite{d20}, the mapping between a face template and a pre-defined random codeword is established via a Locality Sensitive Hashing (LSH) algorithm, while \cite{j21} proposes defining this mapping in terms of a linear transformation represented by an orthogonal matrix. In both cases, as for the Fuzzy Commitment and Fuzzy Vault schemes, successful authentication requires that the correct user-specific codeword be determined, such that its cryptographic hash matches the reference hash stored during enrollment. 

The only Non-NN face BTP method that proposes applying a cryptographic hash function directly to the face template, seems to be \cite{p17}. In this approach, a stacked denoising autoencoder is trained to extract face features from different local regions of the face image. Each local feature vector is then quantised and cryptographically hashed, and the set of local hashes constitutes the protected template. The use of \textit{local} features minimises the intra-class variability across different instances of the same person's face image and, therefore, increases the likelihood of achieving an exact match between the local hashes. Nevertheless, exact matches may still be difficult to ensure in practice; indeed, the high recognition accuracy error rates reported in \cite{p17} are indicative of this issue, which is due to the sensitivity of cryptographic hash functions to small changes in the input (as mentioned earlier).   

Instead of cryptographically hashing random, external codewords, the face BTP methods in \cite{a14, d19, a20, dm20, w20, x20} generate \textit{non-cryptographic} hashes or codes from the NN-extracted face features themselves. The idea is to map similar face templates to (approximately) the same code, thereby achieving a sort of ``hashing'' with reduced sensitivity to intra-class variations (compared to cryptographic hashing). The difference between the BTP methods in \cite{a14, d19, a20, dm20, w20, x20} lies in the way in which this hashing is performed. 

The most well-known non-cryptographic hashing method in the general BTP literature is BioHashing \cite{t06}, whereby a biometric template is projected onto a random subspace (defined by a user-specific matrix) and subsequently binarised via a thresholding operation. BioHashing is applied to NN-extracted face features in \cite{a14}. Random projection, which is the first step in BioHashing, is primarily used to add diversity to the resulting hashes, which further separates different users' templates in the protected domain. This separation, in addition to the binarisation operation, helps to deal with natural variations between different face templates from the same user. Random projection is also applied as a first step towards creating a face hash in \cite{d19, x20}, but the subsequent binarisation of the projected face templates is achieved using an instance of LSH in \cite{d19} and a random decision tree classifier in \cite{x20}. In all three of these BTP methods, the comparison between reference and probe hashes is based on Hamming distance. This allows us to tune the decision threshold according to the acceptable amounts of variation between the same user's face templates (i.e., \textit{intra}-class variability) versus different users' face templates (i.e., \textit{inter}-class variability). This is in contrast to cryptographic hash functions, which would require an \textit{exact} match between the reference and probe hashes (i.e., there would be no need for a decision threshold).

Examples of other BTP methods that propose mapping NN-extracted face features to a binary code, include \cite{dm20, w20}. In \cite{dm20}, 100 features are randomly selected from a higher-dimensional feature vector, and subsets of these features are mapped to their nearest-neighbour binary codewords (where the codewords have been pre-determined via vector quantisation on a development set of features). The binary codewords are then concatenated to generate the protected face template. During authentication, the reference and probe binary templates are compared using Hamming distance. The approach towards generating binary face codes in \cite{w20} is quite different. This BTP method begins by binarising the NN-extracted face feature vector (using its mean), followed by permuting the resulting binary feature vector \textit{m} times to create \textit{m} different permuted vectors. Two permuted binary vectors are then randomly selected and XOR-ed to create the ``bio-key''. As a final step, the bio-key is encoded into a Reed-Solomon (error-correcting) codeword. The comparison between reference and probe codes during authentication then relies on an exact match facilitated by the codes' in-built error-correction capabilities, which differs from the Hamming-distance-based comparisons in \cite{a14, d19, x20, dm20}.

Another example of a non-cryptographic face hashing BTP method was proposed in \cite{a20}. This method differs from \cite{a14, d19, x20, dm20, w20} in that the resulting face hash is not binary, but integer-based. The BTP method in \cite{a20} applies the Winner Takes All (WTA) hashing strategy from \cite{c18} to NN-extracted face features. In a nutshell, this involves permuting the feature vector multiple times, selecting the first \textit{k} elements of each permuted vector, and identifying the indices of the maximum value in each set of \textit{k} elements. Next, the prime factors of each index are computed as: $\texttt{PF} = (\texttt{index} + 2) \times \texttt{random positive integer}$. The final code then consists of a set of integers representing the number of prime factors computed for each permuted face template. During authentication, the probe code is subtracted from the reference code in an element-wise fashion, and the fraction of resulting zeros represents the comparison score.

The hashing methods considered thus far have used \textit{either} cryptographic hash functions \textit{or} non-cryptographic hashing techniques. A face BTP method that combines these two hashing approaches was presented in \cite{h21}. More specifically, \cite{h21} proposes a protected face identification system consisting of two sub-systems. The first sub-system compares the probe face template to each of the \textit{N} reference face templates in the database, where all templates are protected by the non-cryptographic Index-of-Max (IoM) hashing method proposed in \cite{h18} (which is a special instance of LSH). The comparison is based on Hamming distance, and the resulting scores are sorted to identify the top \textit{k} matches, which are passed on to the second sub-system. The second sub-system then conducts a 1-to-1 comparison between the fuzzy vaults of the protected probe template and each of the top \textit{k} protected reference templates returned by the first sub-system. A reference template is considered to match the probe template if there is a match between the cryptographic hashes of the polynomial coefficients extracted from their corresponding fuzzy vaults.  

Among the Non-NN face BTP methods considered thus far, hashing appears to be the most popular approach; however, alternatives are proposed in \cite{a19, p19, kh21, kh22}, where the face templates (extracted features) are \textit{transformed} into protected templates with the help of user-specific transformation functions. For example, in \cite{a19}, face features extracted from different face regions (using a CNN) are fused, and the fused feature vector is convolved with a random kernel. A fusion approach is also adopted in \cite{p19}, but this time a DNN-extracted face feature vector is fused with a different person's face feature vector. Specifically, a key is generated from each feature vector, then each key is used to alter the other feature vector via element-wise multiplication, followed by vector addition of the altered feature vectors. An entirely different BTP method, called \textit{PolyProtect}, is proposed in \cite{kh21, kh22}. This method secures face features (``embeddings'') extracted using pre-trained DNNs, by mapping them to protected templates via multivariate polynomials parameterised by random, user-specific coefficients and exponents. Although such feature transformations may resemble non-cryptographic hashing methods on the surface (i.e., both are ``transforms'' in a sense), the main difference is that hashes are of a \textit{fixed length}, while the size of the protected templates generated by feature transformations is usually \textit{not} fixed (but depends on factors like the original template size). Furthermore, the protected templates generated using feature tranformation BTP methods tend to lie in the same (or similar) domain as the original template (e.g., floating-point values), so the same comparison function can often be applied, while hashes tend to be binary and thus usually necessitate the use of a different comparison function to that adopted in the unprotected template domain (e.g., Hamming distance).

Finally, although the majority of existing Non-NN face BTP methods have been applied at \textit{feature-level}, a couple of \textit{image-level} methods have also been investigated, for example \cite{s18, k20}. In \cite{s18}, the input face image is modified via a block scrambling technique, then a CNN is used to classify the protected image (as a particular identity). In \cite{k20}, face images are warped using distortion functions defined by user-specific keys, then the warped images are presented as inputs to pre-trained face recognition models (or NN feature extractors). The investigation in \cite{k20} actually led to the recommendation that this technique \textit{not} be used for face BTP in practice, because for certain warping parameters the DNN models effectively ignore the warping in the protected images as a form of intra-class variability. It would be interesting to extend this investigation to the image scrambling method in \cite{s18}, although at this point we may reason that a similar conclusion would be reached.

Table II summarises the studied Non-NN face BTP methods, as per the above discussion. The methods are categorised into \textit{feature-level} and \textit{image-level} techniques. Feature-level methods are further divided into \textit{HE} (homomorphic encryption), \textit{hashing} (\textit{cryptographic} and \textit{non-cryptographic}), and \textit{feature transformations}.

\begin{table}[!h]
\renewcommand{\arraystretch}{1.2}
\caption{Examples of Non-NN face BTP methods in the literature.}
\centering
\begin{tabular}{|c|c|c|c|c|}
\hline
\multicolumn{4}{|c|}{\textbf{Feature-level}} & \multirow{3}{*}{\textbf{Image-level}} \\
\cline{1-4}
\multirow{2}{*}{\textit{HE}} & \multicolumn{2}{c|}{\textit{Hashing}} & \textit{Feature} & \\
\cline{2-3}
 & Crypto. & Non-crypto. & \textit{transformations} & \\
\hline
\multirow{3}{*}{\cite{m17, b18, b19, j20, d21a, d21, e20, e22, o21}} & \cite{m19, g19, ko20, ko21, r21, r22, p17}, & \multirow{2}{*}{\cite{a14, d19, dm20},} & \cite{a19}, & \multirow{3}{*}{\cite{s18, k20}} \\
 & \cite{d20, j21}, & \multirow{2}{*}{\cite{w20, a20, x20, h21}} & \cite{p19}, & \\
 & \cite{h21} & & \cite{kh21, kh22} & \\
\hline
\end{tabular}
\end{table}

\subsection{NN-learned Face BTP Methods}

Examples of NN-learned face BTP methods in the literature, include: \cite{p15, p16, z19, j18, jc19, j19, r18, z20, t19, s21, c19, m21, p21, c20, l21}. Recall that methods in this category use a neural network to \textit{learn} the BTP algorithm, instead of explicitly formulating it like Non-NN BTP methods. Furthermore, while the majority of Non-NN face BTP methods investigated in Section II.A were found to be \textit{feature-level} algorithms, such a clear preference for feature-level over image-level techniques was not observed among the NN-learned face BTP methods that will be discussed in this section. This is probably because the main attraction of NN-learned BTP methods lies in the fact that the protection algorithm need not be explicitly defined, meaning that allowing the NN to learn the protected templates from images as opposed to features is not necessarily more challenging (unlike for Non-NN image-level BTP methods, which may additionally require the design of a separate feature extraction algorithm for the protected face images).

The NN-learned methods for face template protection in the literature focus on two main approaches. The first approach involves \textit{pre-defining} a user-specific, random binary code, then learning the mapping from the user's face image/features to this code via a DNN. A cryptographic hash of the pre-defined code then serves as the protected face template, and during authentication the probe and reference codes are compared in the hashed domain. The second approach involves training a DNN to learn its \textit{own representation} of a protected template, without forcing it to conform to a pre-defined code. Since the learned representation is likely to exhibit some intra-class variability (i.e., across the same user's enrollment and various authentication attempts), cryptographic hashing is usually not applied. It is interesting to note that these two types of approaches towards NN-learned face BTP are conceptually similar to the two types of hashing approaches (cryptographic versus non-cryptographic) discussed for Non-NN BTP methods in Section II.A. The difference is that the algorithms used to obtain the protected hashes from the unprotected face templates are \textit{learned} for NN-learned BTP methods and \textit{defined} for Non-NN BTP methods.

Examples of NN-learned face BTP methods that \textit{pre-define} a representative binary code (and cryptographically hash it to produce the protected template), include \cite{p15, p16, z19, j18, jc19, j19}. Of these, \cite{p15, p16, z19} adopt an \textit{image-level} approach, whereby the NN is trained in an end-to-end manner to map the input face images to their corresponding, pre-defined codes. Alternatively, \cite{j18, jc19, j19} adopt a \textit{feature-level} approach, where the NN training starts from the extracted face features as opposed to the raw images. Overall, most of these methods build upon the approach proposed in \cite{p15, p16}, which involves assigning a random maximum-entropy binary code to every user of the face recognition system, then training a Convolutional Neural Network (CNN) to map each user's face image to their corresponding code. During training, the input to the CNN is a face image, and the output of the CNN consists of \textit{n} floating-point values produced by \textit{n} sigmoid activation functions. Each of these outputs is compared to the corresponding bit in the pre-defined \textit{n}-bit code, and the training continues until the \textit{n}-valued output is as close as possible to the \textit{n}-bit code. During the recognition stage, the input face image is passed through the trained CNN to produce the \textit{n} sigmoid outputs, which are then binarised (by setting values above 0.5 to 1 and the rest to 0) to generate the corresponding \textit{n}-bit binary code. The trained CNN should be able to deal with natural variations across different face images acquired from the same user, producing \textit{exactly the same code} during enrollment and each authentication attempt. The binary codes obtained during enrollment and authentication are cryptographically hashed, and comparison is based on an exact match between the two hashes. A similar approach is adopted in \cite{z19}, except that the user-specific binary codes are encoded into error-correcting codes (LDPC) during training, such that the CNN learns the mapping between a user's face image and their corresponding LDPC code. During authentication, the LDPC code obtained for the probe image is decoded to recover the underlying binary code, and the cryptographic hashes of the probe and reference binary codes are compared. This approach aims to improve the CNN's robustness to intra-class variations with the help of the LDPC codes' error-correction capabilities.

Another approach towards improving the BTP method in \cite{p15, p16} was proposed in \cite{j18}. This work adopts a deeper CNN consisting of two components: a pre-trained DNN that is used to extract face features from the input face images, followed by a second NN that maps the extracted face features to the user's pre-defined binary code. This BTP method may thus be considered as a \textit{feature-level} method, since network training does not involve changes to the pre-trained feature extraction network. Indeed, the main idea behind this work is to take advantage of the robust feature extractor to improve the CNN's ability to deal with both intra-class and inter-class variations among the input face images, thereby increasing the recognition accuracy in the protected template domain. 

Further improvements to the BTP approach in \cite{p15, p16}, also based on \textit{feature-level} NN training, were proposed in \cite{jc19, j19}. Both suggestions for improvement mainly focus on the addition of extra, user-specific information to incorporate more complexity into the mapping between a user's face features and their pre-defined (pre-hash) code. Specifically, \cite{jc19} builds on top of \cite{j18}, proposing a random projection step following the feature extraction. In this case, the extracted feature vector is projected to a lower-dimensional code (where the projection is defined by a random, user-specific matrix), after which the NN learns to map this projected code to the user's pre-defined code. The random projection step is motivated by its potential to: (i) increase recognition accuracy by removing redundancies in the face feature vector, (ii) improve security due to the added difficulty in recovering a user's feature vector or face image from their protected template, and (iii) enable the revocability of compromised templates by replacing the user's projection matrix. Similar motivations are presented for the BTP method proposed in \cite{j19}, except that this time the additional user-specific information is incorporated through random perturbations (instead of random projection). Specifically, \cite{j19} proposes applying random, user-specific perturbations to the extracted feature vector, then training the NN to map the perturbed feature vector to its pre-defined binary code. The perturbations are stored as a user-specific key, which is used to apply the same perturbations to the probe feature vector during authentication.   

The main advantage of the NN-learned BTP methods that focus on mapping face images/features to pre-defined codes, is that the resulting protected templates (i.e., cryptographic hashes of the pre-defined codes) are fundamentally unrelated to the original (unprotected) face templates. This means that the protected templates should reveal no information about the face templates to which they have been assigned. Consequently, if an adversary were to gain access to the protected templates stored in the face recognition system's database, they should be unable to recover the corresponding face images or features. Of course, this is based on the assumption that access to the parameters of the trained NN would not help the adversary uncover any hints on the link between a face template and its hash; however, such an assumption may not hold in practice, especially if we consider the worst-case scenario of a fully-informed attacker. The main issue with NN-learned face BTP methods that rely on pre-defined outputs, is that the NN would need to be re-trained (either fully or partially) each time a new user wishes to enroll into the face recognition system or when a compromised user needs to be re-enrolled with a new protected template. So, the scalability of such methods in practice is questionable.  

To avoid the need to re-train the NN for each new enrollment, and thereby improve the scalability of the protected face recognition system, it may be better to use a BTP method that learns its \textit{own} representation of a protected face template, as opposed to learning a mapping to a \textit{pre-defined} representation. Examples of BTP methods in this category, include: \cite{r18, z20, t19, s21, c19, m21, p21, c20, l21}. Of these methods, \cite{r18, z20, c20, l21} adopt a \textit{feature-level} approach, where the NN learns a protected template from extracted face features, and \cite{t19, s21, c19, m21, p21} adopt an \textit{image-level} approach, where the NN is trained in an end-to-end manner to generate the protected template from the face image. A feature-level BTP method can take advantage of robust feature extractors (e.g., pre-trained face recognition models), and it is usually easier to implement/train such methods since the feature extraction part is already taken care of. An image-level approach, on the other hand, would allow for more control over the entire learning process, which could potentially result in more robust protected templates. Although these observations appear intuitive, at this stage it is unclear whether feature-level or image-level NN-learned BTP methods would perform better in practice, especially considering the variety of NN architectures available for each type of approach.

One of the earliest examples of an image-level NN-learned face BTP method, is \cite{t19}. This method proposes a DNN consisting of two components: a Deep Hashing (DH) component and a Neural Network Decoder (NND) component. The DH component is trained to learn an intermediate binary code from the input face image, after which the NND learns to correct errors in this binary code. The error-corrected code is then cryptographically hashed to generate the protected template. Successful authentication relies upon generating exactly the same hash from the probe face image. Since this type of approach learns its \textit{own} code from a face image, as opposed to only learning a mapping to a \textit{pre-defined} code (as in \cite{p15, p16, z19, j18, jc19, j19}), it should technically be possible to enrol new users without the need to re-train the neural network. The reported recognition accuracy in this scenario (referred to as \textit{zero-shot enrollment}), however, is quite low (i.e., 1-2 orders of magnitude worse, depending on the dataset) compared to scenarios where the subject has been seen during the training of the neural network. Furthermore, since the neural network is trained to learn the same code for every face image from the same user, revocability and renewability of compromised protected templates does not seem possible.

The issue of not allowing for template renewability in the event of compromise, would also be present in \cite{s21}, which proposes a face BTP method based on feature fusion. Four different NNs are trained to extract face features from the input face image, and the four sets of learned face features are fused via concatenation. The fused feature vector is then passed on to a fully-connected layer of the ``ensemble network'', where it is classified. Since this method does not require storage of the fused face template, it is claimed that there is no template to be compromised; however, it is conceivable that an adversary with access to the trained NN and its parameters could exploit the classifier to recover a close-enough face image or fused feature vector to match an enrolled identity. In this case, there would be no way for the compromised user to re-enroll into the face recognition system with a new face template.

To allow for the renewability of compromised protected templates, NN-learned BTP methods must incorporate some external (non-face) information into the protected template. This way, if the protected template is compromised, it can be cancelled and replaced by a new protected template generated from the same user's face plus \textit{different} external information. Template renewability was considered in the design of the face BTP methods proposed in \cite{r18, p21, z20, c19, c20, l21, m21}. Most of these methods infuse \textit{user-specific randomness} into the process of learning the protected template, in terms of either a \textit{salting} approach or a \textit{permutation/shuffling} strategy. 

The salting approaches towards incorporating user-specific randomness into the learned protected template, include: \cite{r18, p21, z20}. In \cite{r18}, a pre-trained CNN is used to extract a feature vector from a face image, then the feature vector plus a serial number are passed as inputs to a Recurrent Neural Network (RNN). The output of the RNN, corrected via Reed-Solomon error correction, represents the protected template (referred to as a ``key''). The approach in \cite{p21} uses a modified version of a pre-trained face recognition model to extract a face feature vector, concatenate it with a random key, then pass the result to two fully-connected layers. The NN is trained using a ``secure triplet loss'' formulation, which aims to: (i) minimise the distance between triplet pairs (of face images) originating from the same identity and the same key, (ii) maximise the distance between triplet pairs originating from the same face identity and different keys, as well as from different face identities and different keys, and (iii) control the amount of ``linkability'' between protected templates, where ``unlinkability'' is achieved by ensuring that similar distance values are obtained when comparing protected templates created using different external keys (regardless of whether or not the templates originate from the same identity). As a result of this training procedure, we may reasonably conclude that the renewal of a compromised template, as well as the enrollment of a new user, would not necessarily require re-training of the NN (if it is trained to be sufficiently generalisable in the first place); however, this point is unclear for the method in \cite{r18}, where template renewability is not explicitly considered. 

The final example of the use of salting to incorporate template diversity is \cite{z20}, which begins by using a handcrafted feature extraction algorithm (LBP \cite{a04}) to extract binary codes from face images. Random bits, serving as a ``key'', are then inserted into these codes. The resulting randomised codes are used to train a type of Generative Adversarial Network (GAN) to ``encrypt'' and ``decrypt'' face images. It seems, however, that this method relies on performing reference/probe comparisons in the \textit{decrypted} domain during authentication, which is not secure from a template protection standpoint. In this respect, the BTP method proposed in \cite{z20} appears to present the same limitation as do traditional encryption algorithms, when considering its suitability for securing face templates.

Permutation/shuffling was used to incorporate diversity into the protected templates in \cite{c19, c20, l21, m21}. In \cite{c19}, a pre-trained face recognition DNN is modified and trained to learn a discriminative feature vector from the input face image. This feature vector is then salted by adding Gaussian random noise, and normalised to prevent noise bursts. The NN is further trained to quantise the normalised feature vector, map each quantised segment to the closest binary codeword in a randomly-generated codebook, and binarise and concatenate the codeword indices to generate the final protected template. The entire NN can be trained end-to-end. Although the initial salting operation would incorporate diversity into the protected templates, this step is not mentioned as a motivator for template renewability. Instead, renewability of protected templates is claimed to be achievable by re-shuffling the codewords in the codebook; however, the scalability of this approach before the NN must be re-trained with new codebooks, is unclear.  

In \cite{c20}, the feature vector extracted by a pre-trained face recognition model is permuted via a user-specific permutation matrix, then passed on to a NN that is trained to emulate the handcrafted IoM hashing algorithm \cite{h18}. The protected template is a hash code consisting of the indices of the maximum values resulting from several random projections of the permuted feature vector. A variant of this method is proposed in \cite{l21}, where the permutation step is applied \textit{after} the random projections, instead of to the feature vector as in \cite{c20}. For both methods, template renewability is achievable by changing the user-specific permutation matrix, as this is the source of the non-face randomness. Since the methods were trained and tested on different subjects, it seems that the NN may not need to be re-trained for new/re-enrollments; however, this scalability has not been explicitly evaluated.    

The last example of a NN-learned BTP method that incorporates randomness into the protected face templates, is \cite{m21}. This method trains a ``randomized CNN'' in an end-to-end fashion, incorporating user-specific randomness in three different stages of the process. The first part of the randomized CNN is a feature extraction network, which is trained to extract a discriminative feature vector, $b$, from the face image. Then, the first source of randomness is applied, whereby $b$ is randomly permuted via a user-specific permutation vector, $p$. The permuted feature vector is next split into two parts, $b_A$ and $b_B$. A random binary key, $k$, is generated for the user. This key is secured via a Fuzzy Commitment construct to create a secure sketch, $SS = c \oplus b_B$, where $c$ is an error-correcting code into which $k$ has been encoded.  Then, $k$ is used twice to incorporate additional randomness into the protected template. Firstly, $k$ activates and deactivates randomly selected neurons in RandNet, which is the most important NN component of the randomized CNN. The resulting, randomly-created sub-network takes $b_A$ as its input and outputs a template with partial randomness, $y$. Then, $k$ is used once again, this time to define a ``random permutation-flip'', whereby the elements in $y$ are randomly permuted and the signs of randomly selected elements are flipped. The result is the final protected (and randomised) template, $y_p$, which is stored, along with $SS$ and $p$, in the database during enrollment. During verification, the query protected template, $y^*_p$, is generated using $k$ (decoded from $SS$) and $p$ from the database. Successful verification depends on a high cosine similarity between $y_p$ and $y^*_p$. 

The use of permutations to incorporate user-specific randomness into the protected template in \cite{m21} is by no means a new idea. What makes \cite{m21} stand out from the other NN-learned methods, however, is the idea of randomly modifying the actual architecture of the underlying neural network. The methods in \cite{r18, p21, z20, c19, c20, l21} added randomness either to the extracted feature vector or to the outputs of certain layers of the NN, but \cite{m21} was the first to consider random activation/deactivation of the NN's neurons. The NN in question (i.e., ``randomized CNN'') is trained using a ``randomised triplet loss'', to minimize the distance between templates from the same user generated using the same key and maximize the distance between templates generated using different keys. So, provided that the randomized CNN is well-trained, it should not need to be re-trained for template renewals or new enrollments (especially as the randomized CNN is reported to have been trained and tested on different datasets). There is currently, however, insufficient insight into the scalability of the template renewability process before network re-training may need to be invoked; for example, it would be interesting to investigate how many random sub-network selections would generate sufficiently different protected templates before the father network would need to be modified.

Table III summarises the studied NN-learned face BTP methods. The methods are categorised into \textit{learns pre-defined template} and \textit{learns own template} approaches, depending on what the BTP NN is trained to learn. These categories are further divided into \textit{feature-level} and \textit{image-level} techniques, to indicate whether training of the BTP NN starts from the extracted face features or directly from the face images.  

\begin{table}[!h]
\renewcommand{\arraystretch}{1.1}
\caption{Examples of NN-learned face BTP methods in the literature.}
\centering
\begin{tabular}{|c|c|c|c|}
\hline
\multicolumn{2}{|c|}{\textbf{Learns pre-defined template}} & \multicolumn{2}{c|}{\textbf{Learns own template}} \\
\hline
\textit{Feature-level} & \textit{Image-level} & \textit{Feature-level} & \textit{Image-level} \\
\hline
\cite{j18, jc19, j19} & \cite{p15, p16, z19} & \cite{r18, z20, c20, l21} & \cite{t19, s21, c19, m21, p21} \\
\hline
\end{tabular}
\end{table}

\subsection{Non-NN versus NN-learned Face BTP Methods}

Sections II.A and II.B, respectively, presented examples of Non-NN and NN-learned face BTP methods in the literature. Based on these examples, it is not clear what type of method is preferred among the BTP research community. To get an idea of the current trend, we could consider the number of publications investigating Non-NN versus NN-learned methods within, say, the last 3 years (2018 -- 2021), as illustrated in Fig. 2 for the BTP methods considered in this survey paper.

\begin{figure}[!ht]
\centering
\includegraphics[width=0.8\columnwidth]{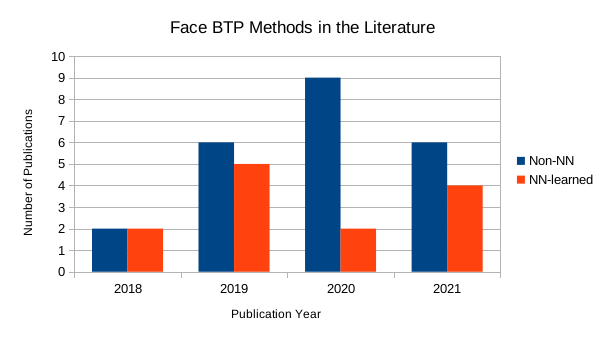}%
\caption{Number of Non-NN and NN-learned face BTP method publications.}
\end{figure}
   
Fig. 2 shows that in 2018 there was an equal focus on Non-NN and NN-learned face BTP methods in the literature. Research into both method types increased from 2018 to 2019, with a slight preference for Non-NN BTP methods. In 2020, there seems to have been a noticeable increase in the preference for Non-NN over NN-learned face BTP methods, probably due to the availability of several pre-trained DNN models that could be used as pre-BTP feature extractors. At the time of writing this paper (about half-way through 2021), interest in Non-NN face BTP methods seems to have dropped a little, while interest in NN-learned methods has started to grow. Although there still seems to be a slight preference for researching Non-NN face BTP methods today, Fig. 2 suggests that this might soon change. In particular, the current gap between Non-NN and NN-learned face BTP methods may close again in 2022, or we might even find that interest in NN-learned BTP methods will surpass the interest in Non-NN methods in the near future. 

Having interpreted Fig. 2 as above, we should emphasize that this is only meant to serve as a rough guide for understanding the research trends in Non-NN versus NN-learned face BTP methods. This is because we cannot guarantee that we have not (accidentally) missed any publications in this survey paper, and moreover the number of publications alone may be inadequate evidence of such trends\footnote{For example, some researchers may work on one BTP idea for a long time before they publish a single paper, whereas others may publish incremental improvements on what is essentially the same method. So, it is not always clear whether a BTP ``trend'' should be defined in terms of the number of \textit{publications} or the number of new BTP method \textit{ideas}.}. Nevertheless, the observed increase in NN-learned face BTP research in 2021 is not surprising, considering the exploding interest in using DNNs to improve the accuracy of face recognition systems in practice (and thus the potential for DNNs to achieve similar performance improvements on the BTP side). Non-NN face BTP methods still seem to be an active area of research, however, so it is important to understand the main conceptual strengths and weaknesses of both types of face BTP methods. The remainder of this section discusses this topic.   

An important strength of Non-NN BTP methods, most of which operate at \textit{feature-level} (see Section II.A), is the flexibility they offer in choosing a feature extractor. This is because feature extraction is performed separately from the BTP step, so as long as the extracted features satisfy the format required by the BTP algorithm, there is some freedom in the choice of feature extraction method. This is particularly important considering the fact that most existing Non-NN face BTP methods rely on face features extracted by pre-trained face recognition DNN models, which have been shown to serve as excellent feature extractors. Since there exist different types of pre-trained models that produce feature vectors of a similar format (i.e., a vector of floating-point values), this means that any Non-NN BTP method that can work with such features can be applied to features extracted using \textit{any} pre-trained face recognition model. This flexibility would be important in practice, where we may imagine the need to integrate a BTP method into an existing face recognition system with an in-built (likely DNN-based) feature extractor.

Although some NN-learned BTP methods were also found to rely on face features extracted by pre-trained face recognition DNN models (see Section II.B), it is difficult to form general conclusions on how sensitive the trained BTP NN would be to changes in the feature extractor (or whether the network would need to be re-trained). This is especially an issue for NN-learned BTP methods that are trained end-to-end, where the part of the NN used to extract features from the input face images is even more intertwined with the part of the NN that serves as the BTP algorithm. Consequently, changing the feature extractor in an end-to-end trained BTP NN would most probably necessitate the re-training of the NN. We may thus conclude that NN-learned BTP methods would be useful in practice in situations where we have greater freedom in the construction of the face recognition system (e.g., choosing the feature extractor, having the option/time/resources to re-train the BTP NN if necessary, etc). In situations where the BTP method must serve as an add-on in a pre-determined face recognition pipeline, however, NN-learned BTP methods may lack the flexibility of Non-NN BTP methods.

The main strength of NN-learned face BTP methods is that they offer the potential for complex (i.e., highly non-linear) protection algorithms that need not be explicitly defined, unlike Non-NN BTP methods. This is an especially attractive aspect of NN-learned BTP methods that adopt an end-to-end learning approach, since we allow the NN to have more freedom in learning how to transform a face image to a protected face template. With this freedom come consequences, however, because the more control we relinquish to the NN, the less certain we become about how exactly the NN is learning what it is learning. As a result, NN-learned BTP methods, particularly those that have been trained end-to-end, tend to be difficult to evaluate. For example, it has already been shown that unprotected DNN-based face recognition models are vulnerable to a \textit{template inversion attack}, whereby a close approximation of the original face image can be obtained from its learned embedding (template) (e.g., \cite{zs16, c17, mc18, os22}). It is thus conceivable that the \textit{protected} face templates learned by NN-learned BTP algorithms could be inverted by similar means to recover the \textit{unprotected} templates and/or the underlying face images; however, such analysis is currently lacking in the face BTP literature. Furthermore, there exists no formal investigation into what information about the unprotected face template is leaked in \textit{different layers} of an NN-learned face BTP model (assuming that an adversary has full access to this model, including its architecture and learned parameters), and how this may \textit{reduce} the amount of protection offered to the face templates. So, while NN-learned BTP methods do have the potential to be more complex, this complexity may come at the price of too much ignorance about how the method works and, therefore, how well it serves to protect the face templates. This is in contrast to Non-NN BTP methods, where the more precise definition of the protection algorithm allows for clearer insight into appropriate evaluation strategies. 

Having commented on the difficulty of evaluating NN-learned BTP methods, we must note that it is by no means impossible to understand the workings of a NN; however, such an undertaking would require a greater level of involvement by the NN's designer. For example, the designer could carefully formulate architectures and loss functions that would more explicitly train the NN to generate protected face templates with certain desirable characteristics, which could both generate stronger templates and make it easier to evaluate the BTP method's robustness. Of course, placing more restrictions on the NN may reduce the attractiveness of using NNs to learn BTP algorithms, since this removes some of the NN's freedom to learn how it will, which forces the BTP method to adopt a more defined algorithm (closer to a Non-NN method). 

Considering these observations and comparisons between Non-NN and NN-learned face BTP methods, it is evident that each type of approach has its own strengths and weaknesses, so there is currently no clear ``winner''; rather, it seems that the two approaches could be combined to build on each other's strengths and reduce each other's weaknesses. This is especially true considering the fact that many of the NN-learned BTP methods were found to be conceptually similar to, or tried to emulate via a DNN, certain Non-NN BTP methods (e.g., see the discussion on cryptographic/non-cryptographic and pre-defined/learned templates in Sections II.A and II.B, respectively). So, it is reasonable to predict that, as this field of research evolves, we will witness a sort of amalgamation between Non-NN and NN-learned BTP methods. This would help to combine the advantages of a more precise BTP algorithm definition with the benefits of using a DNN for greater algorithm complexity, to generate more robust protected templates whose efficacy can be clearly evaluated.

\section{Face BTP Evaluation Techniques}

It is generally agreed upon that an ideal BTP method should satisfy the following three criteria:

\begin{enumerate}

	\item \textbf{Recognition accuracy:} The incorporation of the protection method into a biometric recognition system should not result in a (significant) degradation of the system's recognition accuracy. 
	
	\item \textbf{Irreversibility:} It should be impossible (or computationally infeasible) to recover the original biometric template (features or image) from the protected template.
	
	\item \textbf{Renewability/Unlinkability:} It should be possible to generate multiple \textit{diverse} protected templates from the same subject's original template(s), such that the protected templates \textit{cannot be linked} to the same identity. This would allow for the \textit{cancellation/revocation} and subsequent \textit{renewal} (replacement) of compromised templates, as well as the use of the same biometric characteristic across multiple applications, without the risk of cross-matching the protected templates.
	
\end{enumerate}
	
This section investigates which of these criteria were used to evaluate the face BTP methods discussed in Section II, and the techniques employed to perform the evaluations. Sections III.B, III.C, and III.D present our findings for the ``recognition accuracy'', ``irreversibility'', and ``renewability/unlinkability'' criteria, respectively. Before that, Section III.A considers the reproducibility of the studied BTP works, in terms of the public availability of: the face datasets used to train/test the proposed methods, pre-trained DNN models or known DNN architectures used to implement/evaluate the protected face recognition systems, and the code that ties together the BTP method implementations and their corresponding evaluations.

\subsection{Reproducibility}

A face BTP method's ability to fulfil the aforementioned criteria, especially ``recognition accuracy'', is normally evaluated within the context of a particular face recognition system. Depending on the system's intended operational scenario, the evaluation is further constrained to certain types of face images. So, the evaluation will be influenced by the system into which the BTP method is incorporated and the face images used to test the system. Although many components go into the making of a face recognition system and the setting up of experiments to evaluate it, a full treatise of this topic is outside the scope of this paper. For our purposes, we are interested in the following two aspects of the set-up used to evaluate the face BTP methods discussed in Section II: (i) what face datasets were used, and (ii) what pre-trained DNN models or known architectures were employed within the system.

The ``reproducibility'' of a scientific work refers to the ability of other researchers to replicate (and thereby externally validate) the reported findings (e.g., see \cite{v09}). The most reliable way of ensuring reproducibility is to make the work (e.g., code, experimental protocols, etc.) publicly available, since this would: (i) encourage others to build upon the work without being required to implement the method(s) themselves, and (ii) accurately verify the reported findings without fear of misrepresenting the method(s) or experimental procedure(s). Although it is not easy to ensure 100\% reproducibility even if everything is made public (e.g., some code may only work in certain computing environments), scientific progress favours a fair attempt at reproducibility over no attempt. 

This section investigates the reproducibility of the face BTP works studied in this paper, in terms of the aforementioned experimental set-up. Specifically, we explore the public availability of the employed face datasets and  DNN models/architectures, as well as the code used to implement and evaluate the proposed BTP methods. Tables IV and V present our findings on the employed face datasets, and Table VI provides information on the adopted DNN models/architectures. Finally, Table VII summarises the reproducibility of the face BTP methods and their evaluations, in terms of the public availability of the datasets, DNNs, and code repositories dedicated to reproducing the presented work.


\begin{table*}[!ht]
\renewcommand{\arraystretch}{1.2}
\caption{Face datasets used for evaluating the face BTP methods from Section II.  \textbf{NB:} Train datasets were used to train, fine-tune, or optimise the parameters of, any part of the protected system (e.g., feature extractor, BTP algorithm, classifier). Test datasets were used to report the final evaluation results for the BTP method.} 
\centering
\begin{tabular}{|>{\centering\arraybackslash}p{1.55cm}|>{\centering\arraybackslash}p{1.1cm}|>{\centering\arraybackslash}p{6.9cm}|>{\centering\arraybackslash}p{6.9cm}|}
\hline
\textbf{Method type} & \textbf{Reference} & \textbf{Train dataset} & \textbf{Test dataset} \\
\hline

\multirow{26}{*}{Non-NN} & \cite{m17} & CASIA-WebFace & LFW $\bullet$ Faces94 \\

 \cline{2-4}
 
 & \cite{b18} & -- & LFW $\bullet$ IJB-A $\bullet$ IJB-B $\bullet$ CASIA-WebFace \\
 
 \cline{2-4}
 
 & \cite{b19} & -- & FERET \\
 
 \cline{2-4}
 
 & \cite{j20} & -- & LFW $\bullet$ FEI $\bullet$ Georgia Tech \\
 
 \cline{2-4}
 
 & \cite{d21a, d21} & -- & MORPH \\

 \cline{2-4}
 
 & \cite{e20, e22} & CASIA-WebFace & MegaFace \\
 
 \cline{2-4}
 
 & \cite{o21} & \multicolumn{2}{c|}{Color FERET $\bullet$ FEI $\bullet$ LFW} \\
 
 \cline{2-4}
 
 & \cite{p17} & \multicolumn{2}{c|}{CMU Multi-PIE} \\
 
 \cline{2-4}
 
 & \cite{m19} & MSCeleb-1M $\bullet$ UMDFaces & IJB-A $\bullet$ IJB-C \\
 
 \cline{2-4}
 
 & \cite{ko20, ko21} & VGGFace2 & LFW \\
 
 \cline{2-4}
 
 & \cite{g19} & -- & LFW \\
 
 \cline{2-4}
 
 & \cite{r21, r22} & \multicolumn{2}{c|}{Color FERET $\bullet$ FRGC v2.0} \\
 
 \cline{2-4}
 
 & \cite{a14} & \multicolumn{2}{c|}{AT\&T / ORL $\bullet$ FERET $\bullet$ CMU PIE $\bullet$ M2VTS} \\
 
 \cline{2-4}
 
 & \cite{a19} & \multicolumn{2}{c|}{PaSC $\bullet$ LFW $\bullet$ FERET} \\
 
 \cline{2-4}
 
 & \cite{dm20} & -- & LFW \\
 
 \cline{2-4}
 
 & \cite{s18} & \multicolumn{2}{c|}{AT\&T / ORL} \\
 
 \cline{2-4}
 
 & \cite{k20} & -- & LFW \\
 
 \cline{2-4}
 
 & \cite{d19} & VGGFace2 $\bullet$ LFW & LFW \\
 
 \cline{2-4}
 
 & \cite{d20} & \multicolumn{2}{c|}{CMU PIE $\bullet$ FEI $\bullet$ Color FERET} \\
 
 \cline{2-4}
 
 & \cite{w20} & -- & LFW $\bullet$ AT\&T / ORL \\
 
 \cline{2-4}
 
 & \cite{a20} & -- & FEI $\bullet$ Georgia Tech \\
 
 \cline{2-4}
 
 & \cite{x20} & \multicolumn{2}{c|}{Faces94 $\bullet$ NVIE} \\
 
 \cline{2-4}
 
 & \cite{j21} & -- & CMU Multi-PIE $\bullet$ FEI $\bullet$ Color FERET \\
 
 \cline{2-4}
 
 & \cite{h21} & -- & LFW $\bullet$ VGGFace2 $\bullet$ IJB-C \\
 
 \cline{2-4}
 
 & \cite{p19} & \multicolumn{2}{c|}{Caltech Faces 1999 $\bullet$ Georgia Tech $\bullet$ 10k US Adult Faces} \\
 
 \cline{2-4}
 
 & \cite{kh21, kh22} & -- & MOBIO \\
 
\hline

\multirow{14}{*}{NN-learned} & \cite{p15, p16} & \multicolumn{2}{c|}{CMU PIE $\bullet$ Extended Yale Face Database B $\bullet$ CMU Multi-PIE} \\

 \cline{2-4}
 
 & \cite{z19} & \multicolumn{2}{c|}{CMU PIE $\bullet$ Extended Yale Face Database B} \\
 
 \cline{2-4}
 
 & \cite{j18} & \multicolumn{2}{c|}{CMU PIE $\bullet$ FEI $\bullet$ Color FERET} \\
 
 \cline{2-4}
 
 & \cite{jc19} & \multicolumn{2}{c|}{CMU PIE $\bullet$ FEI $\bullet$ Color FERET} \\
 
 \cline{2-4}
 
 & \cite{j19} & \multicolumn{2}{c|}{CMU PIE $\bullet$ FEI $\bullet$ Color FERET} \\
 
 \cline{2-4}
 
 & \cite{r18} & MS-Celeb-1M $\bullet$ CASIA-WebFace $\bullet$ AT\&T / ORL $\bullet$ Own & CASIA-WebFace $\bullet$ AT\&T / ORL $\bullet$ Own \\
 
 \cline{2-4}
 
 & \cite{z20} & \multicolumn{2}{c|}{LFW} \\
 
 \cline{2-4}
 
 & \cite{t19} & \multicolumn{2}{c|}{CMU PIE $\bullet$ Extended Yale Face Database B $\bullet$ CMU Multi-PIE $\bullet$ WVU Multimodal} \\
 
 \cline{2-4}
 
 & \cite{s21} & \multicolumn{2}{c|}{VGGFace2 $\bullet$ MegaFace} \\
 
 \cline{2-4}
 
 & \cite{c19} & \multicolumn{2}{c|}{YouTube Faces $\bullet$ FaceScrub} \\
 
 \cline{2-4}
 
 & \cite{m21} & VGGFace2 $\bullet$ MS-Celeb-1M & FRGC v2.0 $\bullet$ CFP $\bullet$ IJB-A \\
 
 \cline{2-4}
 
 & \cite{p21} & \multicolumn{2}{c|}{YouTube Faces} \\
 
 \cline{2-4}
 
 & \cite{c20} & \multicolumn{2}{c|}{LFW} \\
 
 \cline{2-4}
 
 & \cite{l21} & YouTube Faces $\bullet$ FaceScrub & LFW $\bullet$ YouTube Faces $\bullet$ FaceScrub \\
 
\hline
\end{tabular}
\end{table*}

Considering Table IV, it is interesting to note that a ``train'' dataset was used in only about \textit{half} of the Non-NN BTP works, compared to \textit{all} of the NN-learned BTP works. This makes sense, because Non-NN face BTP methods tend to employ a neural network only as a feature extractor, meaning that they are more likely to use a pre-trained DNN model for this purpose. On the other hand, for NN-learned BTP methods, a neural network's purpose is to learn the BTP algorithm itself, meaning that training is a necessary step. This is true for at least the BTP part, but possibly for feature extraction as well, especially if the neural network is trained end-to-end.

For the Non-NN BTP works that \textit{did} perform some training, in most cases the motivator was one of the following two reasons. The first reason was to train a reliable NN-based feature extractor (by either fine-tuning an adopted/modified DNN model/architecture, or by training a proprietary DNN), so that a handcrafted BTP algorithm could be applied to the extracted face features. This approach was used in \cite{m17, p17, m19, a14, a19}. Similarly, \cite{e20, e22} trained a DNN to reduce a face feature vector to its ``intrinsic'' dimensionality before encrypting it (via HE), while \cite{s18} and \cite{p19} trained a CNN and an SVM, respectively, to classify already-protected face templates. The second common reason for using a train dataset among Non-NN face BTP methods, was to optimise certain parameters of handcrafted BTP algorithms (which could then be applied to the test dataset) or to train machine-learning BTP methods, but \textit{not} to train a \textit{neural network}. This was the case in \cite{r21, r22, d19, d20, x20}. Similarly, although face templates were protected via HE in \cite{o21}, a separate algorithm (Product Quantisation plus clustering) was trained to extract stable binary codes from the unprotected feature vectors, for indexing purposes in a face identification system. Finally, quite a different motivation for a train dataset was found in \cite{ko20, ko21}, where a neural network was trained to invert templates protected using the Fuzzy Commitment scheme to recover the unprotected face templates.

Considering NN-learned BTP methods, all of which were necessarily trained in some way, Table IV indicates that most of these works used the \textit{same dataset(s)} for both training and testing. In fact, the majority of these (\cite{p15, p16, z19, j18, jc19, j19, r18, c19}) trained and tested their algorithms on even the \textit{same subjects}, whereby the training set of images was used to enroll the subjects and the test set was used to evaluate the recognition accuracy\footnote{In \cite{r18}, MS-Celeb-1M seems to have been used only to train the feature extractor, but the BTP algorithm was trained/tested on the same subjects of the other 3 datasets. Also, \cite{c19} obtained the positive and negative (same/different-subject) test images from two different datasets.}. For \cite{p15, p16, z19, j18, jc19, j19}, this is because the protected templates were \textit{pre-defined} for each subject enrolled in the face recognition system, so the NN was trained only for those subjects (making re-enrollments impossible without re-training the NN, as explained in Section II.B). Although \cite{r18, c19} did not pre-define the protected templates, they still used the same train/test subjects for their evaluations, which makes it difficult to judge the methods' generalisability to unseen subjects.  

The same train/test datasets were also employed for evaluating the NN-learned BTP methods proposed in \cite{t19, p21, c20}. These works, however, explicitly aim to avoid network re-training for new enrollments, by learning their own representations of protected face templates rather than pre-defining them. So, \cite{t19, p21, c20} all used \textit{different subjects} for training and testing their BTP algorithms. Finally\footnote{For the remaining NN-learned BTP works: The train/test split is not explained in \cite{z20}, whereas \cite{s21} used 80:20 and 90:10 train/test splits within each database. It is unclear if the splits contained different subjects.}, \cite{l21, m21} took a step further towards demonstrating the generalisability of their BTP methods, by training and testing their BTP algorithms not only on different subjects, but on different datasets.


\begin{table*}[]
\renewcommand{\arraystretch}{1.2}
\caption{Face datasets from Table IV, arranged first in the order of ``No. uses'' (train/test), then alphabetically.} 
\centering
\begin{tabular}{|c|c|c|c|c|}
\hline
\textbf{Face dataset} & \textbf{No. uses} & \textbf{Paper} & \textbf{Website} & \textbf{Currently available to other researchers?} \\
\hline

\multirow{2}{*}{LFW} & \multirow{2}{*}{14} & \multirow{2}{*}{\cite{h07}} & \multirow{2}{*}{\url{https://bit.ly/38Bu5jk}} & Yes \\

&  &  &  & (freely available) \\
\hline

\multirow{2}{*}{CMU PIE} & \multirow{2}{*}{8} & \multirow{2}{*}{\cite{s02}} & \multirow{2}{*}{\url{https://bit.ly/3kObkPi}} & Yes \\

&  &  &  & (must contact person mentioned on website) \\
\hline

\multirow{2}{*}{FEI} & \multirow{2}{*}{8} & \multirow{2}{*}{\cite{t10}} & \multirow{2}{*}{\url{https://bit.ly/2WMl0S4}} & Yes \\

&  &  &  & (freely available) \\
\hline

\multirow{2}{*}{Color FERET} & \multirow{2}{*}{7} & \multirow{2}{*}{--} & \multirow{2}{*}{\url{https://bit.ly/3mWqUes}} &  Yes \\

&  &  &  &  (must request access) \\
\hline

\multirow{2}{*}{VGGFace2} & \multirow{2}{*}{5} & \multirow{2}{*}{\cite{x18}} & \multirow{2}{*}{\url{https://bit.ly/3kLWWXH} \textit{(unofficial)}} & No \\

&  &  &  & (official download link disappeared) \\
\hline

\multirow{2}{*}{CASIA-WebFace} & \multirow{2}{*}{4} & \multirow{2}{*}{\cite{y14}} & \multirow{2}{*}{\url{https://bit.ly/38BtHRU} \textit{(unofficial)}} & No \\

&  &  &  & (no reliable download link)  \\
\hline

\multirow{2}{*}{CMU Multi-PIE} & \multirow{2}{*}{4} & \multirow{2}{*}{\cite{g10}} & \multirow{2}{*}{\url{https://bit.ly/2WPeM3W}} & Yes \\

&  &  &  & (must order, costs \$250-1000) \\
\hline

\multirow{2}{*}{AT\&T / ORL} & \multirow{2}{*}{3} & \multirow{2}{*}{\cite{s94}} & \multirow{2}{*}{\url{https://bit.ly/3kMB8ev}} & Yes \\

&  &  &  & (freely available) \\
\hline

\multirow{2}{*}{Extended Yale Face Database B} & \multirow{2}{*}{3} & \multirow{2}{*}{\cite{g01}} & \multirow{2}{*}{\url{https://bit.ly/3gUilgq}} & Yes \\

&  &  &  & (freely available) \\
\hline

\multirow{2}{*}{FERET} & \multirow{2}{*}{3} & \multirow{2}{*}{\cite{p00}} & \multirow{2}{*}{\url{https://bit.ly/3mWqUes}} & No \\

&  &  &  & (replaced by Color FERET) \\
\hline

\multirow{2}{*}{Georgia Tech} & \multirow{2}{*}{3} & \multirow{2}{*}{\cite{n97}} & \multirow{2}{*}{\url{https://bit.ly/3yCGfD3}} & Yes \\

&  &  &  & (freely available) \\
\hline

\multirow{2}{*}{MS-Celeb-1M} & \multirow{2}{*}{3} & \multirow{2}{*}{\cite{z16}} & \multirow{2}{*}{\url{https://bit.ly/38zHYOS} \textit{(unofficial)}} & No \\

&  &  &  & (retracted) \\
\hline

\multirow{2}{*}{YouTube Faces} & \multirow{2}{*}{3} & \multirow{2}{*}{\cite{w11}} & \multirow{2}{*}{\url{https://bit.ly/3zDxREp}} & Yes \\

&  &  &  & (freely available after providing contact details) \\
\hline

\multirow{2}{*}{Faces94} & \multirow{2}{*}{2} & \multirow{2}{*}{--} & \multirow{2}{*}{\url{https://bit.ly/3Br35iO}} & Yes \\

&  &  &  & (freely available) \\
\hline

\multirow{2}{*}{FaceScrub} & \multirow{2}{*}{2} & \multirow{2}{*}{\cite{n14}} & \multirow{2}{*}{\url{https://bit.ly/3zRY96l}} & Yes \\

&  &  &  & (must request access) \\
\hline

\multirow{2}{*}{FRGC v2.0} & \multirow{2}{*}{2} & \multirow{2}{*}{\cite{p05}} & \multirow{2}{*}{\url{https://bit.ly/3yxNOLj}} & Yes \\

&  &  &  & (must request access with signed license agreement) \\
\hline

\multirow{2}{*}{IJB-A} & \multirow{2}{*}{2} & \multirow{2}{*}{\cite{k15}} & \multirow{2}{*}{\url{https://bit.ly/38BwKcx}} & Yes \\

&  &  &  & (must request access) \\
\hline

\multirow{2}{*}{MegaFace} & \multirow{2}{*}{2} & \multirow{2}{*}{\cite{k16}} & \multirow{2}{*}{\url{https://bit.ly/3gUniFZ} \textit{(unofficial)}} & No \\

&  &  &  & (retracted) \\
\hline

\multirow{2}{*}{Caltech Faces 1999} & \multirow{2}{*}{1} & \multirow{2}{*}{--} &  \multirow{2}{*}{\url{https://bit.ly/3B0dCDY}} & Yes \\

&  &  &  & (freely available) \\
\hline

\multirow{2}{*}{CFP} & \multirow{2}{*}{1} & \multirow{2}{*}{\cite{s16}} & \multirow{2}{*}{\url{https://bit.ly/3DDqjEj}} & Yes \\

&  &  &  & (freely available) \\
\hline

\multirow{2}{*}{IJB-B} & \multirow{2}{*}{1} & \multirow{2}{*}{\cite{w17}} & \multirow{2}{*}{\url{https://bit.ly/38BwKcx}} & Yes \\

&  &  &  & (must request access) \\
\hline

\multirow{2}{*}{IJB-C} & \multirow{2}{*}{1} & \multirow{2}{*}{\cite{m18}} & \multirow{2}{*}{\url{https://bit.ly/38BwKcx}} & Yes \\

&  &  &  & (must request access) \\
\hline

\multirow{2}{*}{MOBIO} & \multirow{2}{*}{1} & \multirow{2}{*}{\cite{m12}} & \multirow{2}{*}{\url{https://bit.ly/3k6qj84}} & Yes \\

& & & & (must request access with signed license agreement) \\
\hline

\multirow{2}{*}{MORPH (Academic)} & \multirow{2}{*}{1} & \multirow{2}{*}{\cite{r06}} & \multirow{2}{*}{\url{https://bit.ly/3VDQK4U}} & Yes \\

&  &  &  & (must order, costs \$99) \\
\hline

\multirow{2}{*}{M2VTS} & \multirow{2}{*}{1} & \multirow{2}{*}{\cite{m99}} & \multirow{2}{*}{\url{https://bit.ly/3zErAZk}} & Yes \\

&  &  &  & (must order with signed license agreement, costs \pounds{100}-1000) \\
\hline

\multirow{2}{*}{NVIE} & \multirow{2}{*}{1} & \multirow{2}{*}{\cite{l10}} & \multirow{2}{*}{\url{https://bit.ly/3DEe8Hk}} & Yes \\

&  &  &  & (must request access with signed license agreement) \\
\hline

\multirow{2}{*}{Own} & \multirow{2}{*}{1} & \multirow{2}{*}{\cite{r18}} & \multirow{2}{*}{--} & No \\

&  &  &  & (no information on availability) \\
\hline

\multirow{2}{*}{PaSC} & \multirow{2}{*}{1} & \multirow{2}{*}{\cite{b13}} & \multirow{2}{*}{\url{https://bit.ly/2YoPiv3}} & Yes \\

&  &  &  & (must request access with signed license agreement) \\
\hline

\multirow{2}{*}{UMDFaces} & \multirow{2}{*}{1} & \multirow{2}{*}{\cite{b17}} & \multirow{2}{*}{\url{https://bit.ly/3BxPRki}} & No \\

&  &  &  & (temporarily unavailable) \\
\hline

\multirow{2}{*}{10k US Adult Faces} & \multirow{2}{*}{1} & \multirow{2}{*}{\cite{bi13}} & \multirow{2}{*}{\url{https://bit.ly/3EnIbmO}} & Yes \\

&  &  &  & (must request access with signed license agreement) \\
\hline

\multirow{2}{*}{WVU Multimodal} & \multirow{2}{*}{1} & \multirow{2}{*}{\cite{c07}} & \multirow{2}{*}{\url{https://bit.ly/3DEdYjc}} & Partly \\

&  &  &  & (restricted to US citizens) \\
\hline

\end{tabular}
\end{table*}

Next, we turn our attention to Table V, which ranks the face datasets from Table IV in terms of their popularity (measured by the number of face BTP publications that used each dataset for either training or testing). The most popular dataset was LFW, which is understandable given that this dataset has been commonly used to benchmark DNN-based face recognition systems. LFW consists of face images captured in a variety of unconstrained conditions, so it is best used for evaluating face recognition systems designed to operate in such environments. Having said that, before choosing LFW on account of its popularity, the appropriateness of this dataset for evaluating certain face BTP methods should be carefully considered. This is especially true for BTP algorithms that are intended for use in cooperative verification scenarios, which are not well-represented by LFW. This is because the LFW dataset consists of images pulled from the internet, largely of celebrities performing various actions (e.g., speaking, playing a sport), so they do not resemble the types of images that are likely to be acquired during a deliberate verification attempt. 

The second most widely-used face datasets among the studied face BTP works are CMU PIE and FEI. However, since CMU Multi-PIE is actually an extension of CMU PIE, the combined dataset is more popular than FEI and almost as widely-adopted as LFW. CMU (Multi-)PIE consists of face images acquired in a controlled environment, but where each subject's face images have been captured from multiple viewpoints, under various lighting conditions, while displaying a range of facial expressions. So, this dataset contains lots of deliberate variations, and variation factors that are deemed inappropriate for certain evaluations can be excluded. As a result, CMU (Multi-)PIE is easier to use than LFW for evaluating face BTP methods intended for deployment in cooperative verification scenarios, where the acquired face images may contain certain (reasonable) inconsistencies.

Another interesting observation from Table V is that, of the 31 datasets that are listed, \textit{over half} have been employed by only 1 or 2 of the studied face BTP works. One reason for this could be that these datasets were considered the most suitable for evaluating the corresponding face BTP methods, so they were chosen despite the fact that other researchers did not use the same ones for evaluating their own BTP methods. If this is indeed the case, then the approach is reasonable, although this makes it difficult to fairly compare the performance of different BTP methods. To facilitate this step, it is important that BTP researchers work towards the reproducibility of their method implementations, as discussed earlier.

Considering reproducibility from the point of view of the datasets used to evaluate the studied face BTP methods, Table V indicates that the majority (23 out of 31) of these datasets are currently available to other researchers. Most of the remaining 8 datasets \textit{used to be} publicly available, but have now been either permanently or temporarily retracted (usually for legal reasons). So, although we can never guarantee that a dataset will forever be available, it is encouraging to observe that most of the face BTP methods studied in this paper have been trained/evaluated on datasets that were publicly available at least at the time of their publication. A continuing focus on using public datasets in the BTP research community will help to increase the reproducibility of the published work, thereby fostering greater understanding of the proposed BTP methods.  

Aside from public availability of the employed datasets, the most important facilitator of reproducibility would be the public availability of the code used to implement and evaluate the proposed BTP methods. An important aspect of this concerns the use of publicly available pre-trained DNN models or implementations of known DNN architectures, which is summarised in Table VI for the studied face BTP works. 


\begin{table*}[]
\renewcommand{\arraystretch}{1.05}
\caption{Pre-trained DNN models or known DNN architectures used to implement/evaluate the protected face recognition systems. \newline \textbf{NB:} ``Link to public implementation'' was obtained from either ``reference'' or ``DNN cited paper''.} 
\centering
\begin{tabular}{|c|c|c|c|c|}
\hline
\textbf{Method type} & \textbf{Reference} & \textbf{DNN name} & \textbf{DNN cited paper} & \textbf{Link to public DNN implementation} \\

\hline

\multirow{51}{*}{Non-NN} & \multirow{2}{*}{\cite{m17}} & DeepID & \cite{s14} & {\color{red}\ding{56}} \\

\cline{3-5}

 & & Light CNN-9 & \cite{w15} & \url{https://bit.ly/3kJCKFK} \\

 \cline{2-5}
 
 & \multirow{3}{*}{\cite{b18}} & \multirow{2}{*}{FaceNet} & \multirow{2}{*}{\cite{s15}} & \url{https://bit.ly/3BFc2Fm} \\
 
 & & & & (model: 128-dim. output) \\
 
 \cline{3-5}
 
 & & SphereFace & \cite{l17} & \url{https://bit.ly/3gVqf96} \\
 
 \cline{2-5}
 
 & \cite{b19} & FaceNet & \cite{s15} & {\color{red}\ding{56}} \\
 
 \cline{2-5}
 
 & \cite{j20} & FaceNet & \cite{s15} & {\color{red}\ding{56}} \\
 
 \cline{2-5}
 
 & \multirow{4}{*}{\cite{d21a, d21}} & \multirow{2}{*}{ArcFace} & \multirow{2}{*}{\cite{x19}} & \url{https://bit.ly/2WTus6v} \\
 
 & & & & (model: ``LResNet100E-IR,ArcFace@ms1m-refine-v2'') \\
 
 \cline{3-5}
 
 & & \multirow{2}{*}{CurricularFace} & \multirow{2}{*}{\cite{h20}} & \url{https://bit.ly/3zHPovx} \\
 
 & & & & (model: ``IR101'') \\
 
 \cline{2-5}
 
 & \multirow{3}{*}{\cite{e20, e22}} & \multirow{2}{*}{ArcFace} & \multirow{2}{*}{\cite{x19}} & \url{https://bit.ly/2WTus6v} \\
 
 & & & & (model: 512-dim. output) \\
 
 \cline{3-5}
 
 & & DeepMDS & \cite{gb19} & \url{https://bit.ly/3bk9lBG} \\
 
 \cline{2-5}
 
 & \multirow{2}{*}{\cite{o21}} & \multirow{2}{*}{ArcFace} & \multirow{2}{*}{\cite{x19}} & \url{https://bit.ly/2WTus6v} \\
 
 & & & & (model: ResNet-100) \\
 
 \cline{2-5}
 
 & \cite{p17} & -- & -- & -- \\
 
 \cline{2-5}
 
 & \cite{m19} & ResNeXt & \cite{x17} & \url{https://bit.ly/3kTXJFU} \\
 
 \cline{2-5}
 
 & \multirow{6}{*}{\cite{ko20, ko21}} & \multirow{2}{*}{FaceNet} & \multirow{2}{*}{\cite{s15}} & \url{https://bit.ly/2WJTogQ} \\
 
 & & & & (model: 128-dim. output) \\
 
 \cline{3-5}
 
 & & Dlib & \cite{h16} & \url{https://bit.ly/3h1qq2S} \\
 
 \cline{3-5}
 
 & & Light CNN-29v2 & \cite{wh18} & \url{https://bit.ly/3kJCKFK} \\
 
 \cline{3-5}
 
 & & NbNet & \cite{mc18} & {\color{red}\ding{56}} \\
 
 \cline{3-5}
 
 & & ArcFace & \cite{x19} & \url{https://bit.ly/2WTus6v} \\
 
 \cline{2-5}
 
 & \cite{g19} & VGG-Face & \cite{v15} & \url{https://bit.ly/3tpmGgF} \\
 
 \cline{2-5}
 
 & \multirow{2}{*}{\cite{r21, r22}} & \multirow{2}{*}{ArcFace} & \multirow{2}{*}{\cite{x19}} & \url{https://bit.ly/2WTus6v} \\
 
 & & & & (model: ``LResNet100E-IR,ArcFace@ms1m-refine-v2'') \\
 
 \cline{2-5}
 
 & \cite{a14} & -- & -- & -- \\
 
 \cline{2-5}
 
 & \cite{a19} & -- & -- & -- \\
 
 \cline{2-5}
 
 & \cite{dm20} & VGG-Face & \cite{v15} & \url{https://bit.ly/3tpmGgF} \\
 
 \cline{2-5}
 
 & \cite{s18} & -- & -- & -- \\
 
 \cline{2-5}
 
 & \multirow{3}{*}{\cite{k20}} & ArcFace & \cite{x19} & \url{https://bit.ly/2WTus6v} \\
 
 \cline{3-5}
 
 & & MobileFace & \cite{l18} & {\color{red}\ding{56}} \\
 
 \cline{3-5}
 
 & & ShuffleFace & \cite{md19} & {\color{red}\ding{56}} \\
 
 \cline{2-5}
 
 & \multirow{2}{*}{\cite{d19}} & \multirow{2}{*}{ArcFace} & \multirow{2}{*}{{\color{red}\ding{56}}} & \url{https://bit.ly/2WTus6v} \\
 
 & & & & (model: pre-trained on MS-Celeb-1M) \\
 
 \cline{2-5}
 
 & \multirow{2}{*}{\cite{d20}} & \multirow{2}{*}{FaceNet} & \multirow{2}{*}{\cite{s15}} & \url{https://bit.ly/3BFc2Fm} \\
 
 & & & & (model: 512-dim. output) \\
 
 \cline{2-5}
 
 & \cite{w20} & FaceNet & \cite{s15} & {\color{red}\ding{56}} \\
 
 \cline{2-5}
 
 & \cite{a20} & AlexNet & \cite{k12} & \url{https://bit.ly/38CKSlT} \\
 
 \cline{2-5}
 
 & \multirow{2}{*}{\cite{x20}} & LeNet-5 & \cite{l98} & {\color{red}\ding{56}} \\
 
 \cline{3-5}
 
 & & AlexNet & \cite{k12} & \url{https://bit.ly/38CKSlT} \\
 
 \cline{2-5}
 
 & \multirow{3}{*}{\cite{j21}} & \multirow{2}{*}{ArcFace} & \multirow{2}{*}{\cite{x19}} & \url{https://bit.ly/2WTus6v} \\
 
 & & & & (model: ResNet50, 512-dim. output) \\
 
 \cline{3-5}
 
 & & CosFace & \cite{g18} & {\color{red}\ding{56}} \\
 
 \cline{2-5}
 
 & \multirow{3}{*}{\cite{h21}} & FaceNet & \cite{s15} & {\color{red}\ding{56}} \\
 
 \cline{3-5}
 
 & & \multirow{2}{*}{ArcFace} & \multirow{2}{*}{\cite{x19}} & \url{https://bit.ly/2WTus6v} \\
 
 & & & & (model: pre-trained on MS-Celeb-1M, 256-dim. output) \\
 
 \cline{2-5}
 
 & \multirow{2}{*}{\cite{p19}} & \multirow{2}{*}{FaceNet} & \multirow{2}{*}{\cite{s15}} & \url{https://bit.ly/3BFc2Fm} \\
 
 & & & & (model: Inception-ResNet-v1, 512-dim. output) \\
 
 \cline{2-5}

 & \multirow{2}{*}{\cite{kh21, kh22}} & FaceNet & {\color{red}\ding{56}} & \url{https://bit.ly/3nv3FYV} \\ 
 
 \cline{3-5}
 
 & & Idiap & {\color{red}\ding{56}} & \url{https://bit.ly/3kaLTYQ} \\
 
 \hline

\multirow{17}{*}{NN-learned} & \cite{p15, p16} & -- & -- & -- \\

 \cline{2-5}
 
 & \cite{z19} & CaffeNet & \cite{j14} & \url{https://bit.ly/3BELMLm} \\
 
 \cline{2-5}
 
 & \cite{j18} & VGG-Face & \cite{v15} & \url{https://bit.ly/3tpmGgF} \\
 
 \cline{2-5}
 
 & \cite{jc19} & VGG-Face & \cite{v15} & \url{https://bit.ly/3tpmGgF} \\
 
 \cline{2-5}
 
 & \cite{j19} & VGG-Face & \cite{v15} & \url{https://bit.ly/3tpmGgF} \\
 
 \cline{2-5}
 
 & \cite{r18} & Inception-ResNet-v1 & \cite{s17} & \url{https://bit.ly/3kNNCCJ} \\
 
 \cline{2-5}
 
 & \cite{z20} & ResNet-11 & {\color{red}\ding{56}} & {\color{red}\ding{56}} \\
 
 \cline{2-5}
 
 & \cite{t19} & VGG-19 & \cite{z14} & \url{https://bit.ly/2Yt3x1X} \\
 
 \cline{2-5}
 
 & \multirow{2}{*}{\cite{s21}} & ResNet & \cite{h16} & {\color{red}\ding{56}} \\
 
 \cline{3-5}
 
 & & DenseNet & \cite{h17} & \url{https://bit.ly/38DAFG4} \\
 
 \cline{2-5}
 
 & \cite{c19} & VGG-16 & \cite{z14} & \url{https://bit.ly/2Yt3x1X} \\
 
 \cline{2-5}
 
 & \cite{m21} & ResNet-50 & \cite{h16} & {\color{red}\ding{56}} \\
 
 \cline{2-5}
 
 & \cite{p21} & Inception-ResNet-v1 & \cite{s17} & \url{https://bit.ly/2WI8hQd} \\
 
 \cline{2-5}
 
 & \multirow{2}{*}{\cite{c20}} & \multirow{2}{*}{ArcFace} & \multirow{2}{*}{\cite{x19}} & \url{https://bit.ly/2WTus6v}  \\
 
 & & & & (model: ResNet-50, 512-dim. output) \\
 
 \cline{2-5}
 
 & \multirow{2}{*}{\cite{l21}} & \multirow{2}{*}{ArcFace} & \multirow{2}{*}{\cite{x19}} & \url{https://bit.ly/2WTus6v} \\
 
 & & & & (model: ResNet-50, 512-dim. output) \\
 
\hline
\end{tabular}
\end{table*}

Table VI indicates that ArcFace and FaceNet pre-trained models are the most popular. Apart from their impressive performance, the widespread adoption of ArcFace models may be attributed to the fact that the model repository (\url{https://bit.ly/2WTus6v}) is well-maintained by two of the authors of \cite{x19}. For FaceNet, unfortunately there exists no public implementation of \cite{s15} by its authors, but a number of models inspired by \cite{s15} have been made available by others. The most popular FaceNet model repository appears to be that of David Sandberg (\url{https://bit.ly/3BFc2Fm}), but this is not the only one (e.g., \url{https://bit.ly/2WJTogQ} presents another version). So, when authors do not point to a particular FaceNet implementation (e.g., \cite{b19, j20, w20, h21} in Table VI), the reproducibility of their work is reduced. In general, however, Table VI shows that only about a third of the studied face BTP works did not provide a clear indication of where their adopted DNN models/architectures came from, whereas the remaining two thirds provided either a direct link or sufficient information to locate the adopted DNN (e.g., a citation of the DNN's paper, where a public repository link is provided). 


\begin{table*}[!ht]
\renewcommand{\arraystretch}{1.2}
\caption{Reproducibility of the face BTP methods and evaluations. \textbf{NB:} The order of multiple {\color{olive}\ding{52}}, {\color{red}\ding{56}}, and/or {\color{blue}$\thickapprox$} for ``datasets'' and ``DNNs'' corresponds to the order of these elements in Tables IV and VI, respectively (for a specific reference).} 
\centering
\begin{tabular}{|c|c|>{\centering\arraybackslash}p{2cm}|>{\centering\arraybackslash}p{2cm}|c|c|}
\hline
\multirow{2}{*}{\textbf{Method type}} & \multirow{2}{*}{\textbf{Reference}} & \multicolumn{2}{c|}{\textbf{Are used datasets (still) available?}} & \multirow{2}{*}{\textbf{Link to employed DNN(s)?}} & \multirow{2}{*}{\textbf{Link to paper's public code repository?}} \\

\cline{3-4}

 & & \textbf{Train} & \textbf{Test} & & \\

\hline

\multirow{27}{*}{Non-NN} & \cite{m17} & \multicolumn{1}{c|}{{\color{red}\ding{56}}} & \multicolumn{1}{c|}{{\color{olive}\ding{52}}{\color{olive}\ding{52}}} & {\color{red}\ding{56}}{\color{olive}\ding{52}} & {\color{red}\ding{56}} \\

\cline{2-6}

 & \cite{b18} & -- & \multicolumn{1}{c|}{{\color{olive}\ding{52}}{\color{olive}\ding{52}}{\color{olive}\ding{52}}{\color{red}\ding{56}}} & \multicolumn{1}{c|}{{\color{olive}\ding{52}}{\color{olive}\ding{52}}} & \url{https://bit.ly/3DIPJjR} \\

 \cline{2-6}
 
 & \cite{b19} &  -- & \multicolumn{1}{c|}{{\color{red}\ding{56}}} & \multicolumn{1}{c|}{{\color{red}\ding{56}}} & {\color{red}\ding{56}} \\
 
 \cline{2-6}
 
 & \cite{j20} &  -- & \multicolumn{1}{c|}{{\color{olive}\ding{52}}{\color{olive}\ding{52}}{\color{olive}\ding{52}}} & \multicolumn{1}{c|}{{\color{red}\ding{56}}} & {\color{red}\ding{56}} \\
 
 \cline{2-6}
 
 & \cite{d21a, d21} & -- & \multicolumn{1}{c|}{{\color{olive}\ding{52}}} & \multicolumn{1}{c|}{{\color{olive}\ding{52}}{\color{olive}\ding{52}}} & {\color{red}\ding{56}} \\
 
 \cline{2-6}
 
  & \multirow{2}{*}{\cite{e20, e22}} & \multicolumn{1}{c|}{\multirow{2}{*}{{\color{red}\ding{56}}}} & \multicolumn{1}{c|}{\multirow{2}{*}{{\color{red}\ding{56}}}} & \multicolumn{1}{c|}{\multirow{2}{*}{{\color{olive}\ding{52}}{\color{olive}\ding{52}}}} & \multicolumn{1}{c|}{\multirow{2}{*}{}{\url{https://bit.ly/3bk9lBG}}} \\
  & & & & & (Available in 2022 \cite{e22}, not in 2020 \cite{e20}.) \\
 
 \cline{2-6}
 
 & \cite{o21} & \multicolumn{2}{c|}{{\color{olive}\ding{52}}{\color{olive}\ding{52}}{\color{olive}\ding{52}}} & \multicolumn{1}{c|}{{\color{olive}\ding{52}}} & {\color{red}\ding{56}} \\
 
 \cline{2-6}
 
 & \cite{p17} & \multicolumn{2}{c|}{{\color{olive}\ding{52}}} & -- & {\color{red}\ding{56}} \\
 
 \cline{2-6}
 
 & \cite{m19} & \multicolumn{1}{c|}{{\color{red}\ding{56}}{\color{red}\ding{56}}} & \multicolumn{1}{c|}{{\color{olive}\ding{52}}{\color{olive}\ding{52}}} & {\color{olive}\ding{52}} & {\color{red}\ding{56}} \\
  
 \cline{2-6}
  
 & \cite{ko20, ko21} & \multicolumn{1}{c|}{{\color{red}\ding{56}}} & \multicolumn{1}{c|}{{\color{olive}\ding{52}}} & \multicolumn{1}{c|}{{\color{olive}\ding{52}}{\color{olive}\ding{52}}{\color{olive}\ding{52}}{\color{red}\ding{56}}{\color{olive}\ding{52}}} & {\color{red}\ding{56}} \\
 
 \cline{2-6}
 
 & \cite{g19} & -- & \multicolumn{1}{c|}{{\color{olive}\ding{52}}} & {\color{olive}\ding{52}} & {\color{red}\ding{56}} \\
 
 \cline{2-6}
 
 & \cite{r21, r22} & \multicolumn{2}{c|}{{\color{olive}\ding{52}}{\color{olive}\ding{52}}} & {\color{olive}\ding{52}} & {\color{red}\ding{56}} \\
 
 \cline{2-6}
 
 & \cite{a14} & \multicolumn{2}{c|}{{\color{olive}\ding{52}}{\color{red}\ding{56}}{\color{olive}\ding{52}}{\color{olive}\ding{52}}} & -- & {\color{red}\ding{56}} \\

 \cline{2-6}
 
 & \cite{a19} & \multicolumn{2}{c|}{{\color{olive}\ding{52}}{\color{olive}\ding{52}}{\color{red}\ding{56}}} & -- & {\color{red}\ding{56}} \\
 
 \cline{2-6}
 
 & \cite{dm20} & -- & \multicolumn{1}{c|}{{\color{olive}\ding{52}}} & \multicolumn{1}{c|}{{\color{olive}\ding{52}}} & \url{https://bit.ly/2YqeGAr} \\
 
 \cline{2-6}
 
 & \cite{s18} & \multicolumn{2}{c|}{{\color{olive}\ding{52}}} & -- & {\color{red}\ding{56}} \\
 
 \cline{2-6}
 
 & \cite{k20} & -- & \multicolumn{1}{c|}{{\color{olive}\ding{52}}} & \multicolumn{1}{c|}{{\color{olive}\ding{52}}{\color{red}\ding{56}}{\color{red}\ding{56}}{\color{red}\ding{56}}} & {\color{red}\ding{56}} \\
 
 \cline{2-6}
 
 & \cite{d19} & \multicolumn{1}{c|}{{\color{red}\ding{56}}{\color{olive}\ding{52}}} & \multicolumn{1}{c|}{{\color{olive}\ding{52}}} & {\color{olive}\ding{52}} & \url{https://bit.ly/3VHY5jX} \\
 
 \cline{2-6}
 
 & \cite{d20} & \multicolumn{2}{c|}{{\color{olive}\ding{52}}{\color{olive}\ding{52}}{\color{olive}\ding{52}}} & {\color{olive}\ding{52}} & {\color{red}\ding{56}} \\
 
 \cline{2-6}
 
 & \cite{w20} & -- & \multicolumn{1}{c|}{{\color{olive}\ding{52}}{\color{olive}\ding{52}}} & {\color{red}\ding{56}} & {\color{red}\ding{56}} \\
 
 \cline{2-6}
 
 & \cite{a20} & -- & \multicolumn{1}{c|}{{\color{olive}\ding{52}}{\color{olive}\ding{52}}} & {\color{olive}\ding{52}} & {\color{red}\ding{56}} \\
 
 \cline{2-6}
 
 & \cite{x20} & \multicolumn{2}{c|}{{\color{olive}\ding{52}}{\color{olive}\ding{52}}} & {\color{red}\ding{56}}{\color{olive}\ding{52}} & {\color{red}\ding{56}} \\
 
 \cline{2-6}
 
 & \cite{j21} & -- & \multicolumn{1}{c|}{{\color{olive}\ding{52}}{\color{olive}\ding{52}}{\color{olive}\ding{52}}} & {\color{olive}\ding{52}}{\color{red}\ding{56}} & {\color{red}\ding{56}} \\
 
 \cline{2-6}
 
 & \cite{h21} & -- & \multicolumn{1}{c|}{{\color{olive}\ding{52}}{\color{red}\ding{56}}{\color{olive}\ding{52}}} & {\color{red}\ding{56}}{\color{olive}\ding{52}} & {\color{red}\ding{56}} \\
 
 \cline{2-6}
 
  & \cite{p19} & \multicolumn{2}{c|}{{\color{olive}\ding{52}}{\color{olive}\ding{52}}{\color{olive}\ding{52}}} & {\color{olive}\ding{52}} & {\color{red}\ding{56}} \\
 
 \cline{2-6}
 
 & \cite{kh21, kh22} & -- & \multicolumn{1}{c|}{{\color{olive}\ding{52}}} & {\color{olive}\ding{52}} & \url{https://bit.ly/3uBw5C6} \\
 
\hline

\multirow{14}{*}{NN-learned} & \cite{p15, p16} & \multicolumn{2}{c|}{{\color{olive}\ding{52}}{\color{olive}\ding{52}}{\color{olive}\ding{52}}} & -- & {\color{red}\ding{56}} \\

 \cline{2-6}
 
 & \cite{z19} & \multicolumn{2}{c|}{{\color{olive}\ding{52}}{\color{olive}\ding{52}}} & {\color{olive}\ding{52}} & \url{https://bit.ly/3mZA8GH} \\
 
 \cline{2-6}
 
 & \cite{j18} & \multicolumn{2}{c|}{{\color{olive}\ding{52}}{\color{olive}\ding{52}}{\color{olive}\ding{52}}} & {\color{olive}\ding{52}} & {\color{red}\ding{56}} \\
 
 \cline{2-6}
 
 & \cite{jc19} & \multicolumn{2}{c|}{{\color{olive}\ding{52}}{\color{olive}\ding{52}}{\color{olive}\ding{52}}} & {\color{olive}\ding{52}} & {\color{red}\ding{56}} \\
 
 \cline{2-6}
 
 & \cite{j19} & \multicolumn{2}{c|}{{\color{olive}\ding{52}}{\color{olive}\ding{52}}{\color{olive}\ding{52}}} & {\color{olive}\ding{52}} & {\color{red}\ding{56}} \\
 
 \cline{2-6}
 
 & \cite{r18} & \multicolumn{1}{c|}{{\color{red}\ding{56}}} & \multicolumn{1}{c|}{{\color{red}\ding{56}}{\color{olive}\ding{52}}{\color{red}\ding{56}}} & {\color{olive}\ding{52}} & {\color{red}\ding{56}} \\
 
 \cline{2-6}
 
 & \cite{z20} & \multicolumn{2}{c|}{{\color{olive}\ding{52}}} & {\color{red}\ding{56}} & {\color{red}\ding{56}} \\
 
 \cline{2-6}
 
 & \cite{t19} & \multicolumn{2}{c|}{{\color{olive}\ding{52}}{\color{olive}\ding{52}}{\color{olive}\ding{52}}{\color{blue}$\thickapprox$}} & {\color{olive}\ding{52}} & {\color{red}\ding{56}} \\
 
 \cline{2-6}
 
 & \cite{s21} & \multicolumn{2}{c|}{{\color{red}\ding{56}}{\color{red}\ding{56}}} & {\color{red}\ding{56}}{\color{olive}\ding{52}} & {\color{red}\ding{56}} \\
 
 \cline{2-6}
 
 & \cite{c19} & \multicolumn{2}{c|}{{\color{olive}\ding{52}}{\color{olive}\ding{52}}} & {\color{olive}\ding{52}} & {\color{red}\ding{56}} \\
 
 \cline{2-6}
 
 & \cite{m21} & \multicolumn{1}{c|}{{\color{red}\ding{56}}{\color{red}\ding{56}}} & \multicolumn{1}{c|}{{\color{olive}\ding{52}}{\color{olive}\ding{52}}{\color{olive}\ding{52}}} & {\color{red}\ding{56}} & {\color{red}\ding{56}} \\
 
 \cline{2-6}
 
 & \cite{p21} & \multicolumn{2}{c|}{{\color{olive}\ding{52}}} & {\color{olive}\ding{52}} & \url{https://bit.ly/3gZ6eOU} \\
 
 \cline{2-6}
 
 & \cite{c20} & \multicolumn{2}{c|}{{\color{olive}\ding{52}}} & {\color{olive}\ding{52}} & {\color{red}\ding{56}} \\
 
 \cline{2-6}
 
 & \cite{l21} & \multicolumn{1}{c|}{{\color{olive}\ding{52}}{\color{olive}\ding{52}}} & \multicolumn{1}{c|}{{\color{olive}\ding{52}}{\color{olive}\ding{52}}{\color{olive}\ding{52}}} & {\color{olive}\ding{52}} & {\color{red}\ding{56}} \\
 
\hline
\end{tabular}
\end{table*}

Finally, the most important way of ensuring reproducibility is to make public the code that ties together the entire implementation of the BTP algorithm (including any adopted DNN models/architectures), as well as the procedures used to evaluate it. Table VII shows that only a handful of the studied face BTP works provided a link to a public code repository intended for reproducing their work (\cite{b18, e22, dm20, d19, kh21, kh22, z19, p21}). Although we did not test all this code ourselves and cannot, therefore, guarantee that each link points to fully reproducible work, we still consider the presence of a public code repository a deliberate attempt at reproducibility. Note that some papers provided links to \textit{other} open-source code that they used for certain parts of their work, but not to a code repository specifically dedicated to reproducing their proposed methods. For example, \cite{ko20, ko21} provided a link to the adopted Fuzzy Commitment (FC) implementation, but not to code using the FC to launch the template reconstruction attack that was the whole point of this paper. Similarly, \cite{o21} provided a link to the SEAL library used for the HE component of this work, but not to code implementing the actual search/indexing method for retrieving the protected face templates in a face identification system, which was the main contribution of this paper. Although such individual links are useful for enabling others to implement certain \textit{components} of the proposed methods/experiments, they were not mentioned in Table VII because they do not allow the work as a whole to be reproduced, i.e., the \textit{combination} of these individual ``components'' is left open to interpretation.

Taking into account the public availability of the papers' code repositories, any adopted DNN models/architectures, and the employed datasets, Table VII implies that, at this stage, it would be difficult to faithfully reproduce most of the existing face BTP methods. Of course, there is always the possibility of private dataset/model/code exchanges between individual researchers; however, this does not benefit the BTP research community as a whole, which would only be achieved if all work was made freely and equally available to all researchers.

Although the reproducibility of a BTP method aids in its visibility and further development, we cannot launch into reproducing a piece of work before understanding what has been proposed. Furthermore, even though the availability of all BTP work to the entire research community would be a commendable achievement, this may eventually lead to information overload. In this case, researchers will have to choose which BTP methods to build upon or (re-)evaluate, meaning that an understanding of the methods and their evaluation procedures in advance would be helpful. Section II of this paper discussed the types of face BTP methods proposed thus far, and the current section investigated the reproducibility of these works. Next, Sections III.B, III.C, and III.D explore the techniques that were used to evaluate the ``recognition accuracy'', ``irreversibility'', and ``renewability/unlinkability'' criteria, respectively, for the existing face BTP methods.

\subsection{Recognition Accuracy}

A BTP method is considered to satisfy the ``recognition accuracy'' criterion if it can be incorporated into a biometric system without (significantly) degrading its recognition accuracy. In other words, the ability of the biometric system to recognise the enrolled identities should be approximately the same regardless of whether the identities are represented in terms of their protected or unprotected biometric templates. This section considers which face BTP methods from Section II were evaluated for their ability to satisfy this criterion and what metrics/plots were used to conduct the evaluation.

Of the face BTP works discussed in Section II, all but \cite{ko20, ko21} presented an evaluation of the protected system's recognition accuracy. Evaluation of this criterion was not deemed necessary in \cite{ko20, ko21}, since the aim was to show that DNN-extracted face templates protected using the Fuzzy Commitment scheme can be inverted using a reconstruction attack (meaning that this BTP method satisfies neither the irreversibility nor the renewability/unlinkability criterion). Considering the publications that \textit{did} evaluate the recognition accuracy of their proposed BTP methods, the adopted evaluation metrics/plots were usually the same as those used to evaluate the recognition accuracy of a standard (unprotected) biometric system. This is to be expected, since a BTP algorithm must operate \textit{within} a biometric system, meaning that the recognition accuracy evaluation should reflect the system's ability to achieve its underlying aim (i.e., to recognise the enrolled users).

Table VIII summarises the metrics and plots used to evaluate the recognition accuracy of the studied face BTP methods. Note that other metrics/plots may have been adopted for some evaluations, but Table VIII presents only those used to report the \textit{final} recognition accuracy (not those used in intermediate processing stages, like for fine-tuning system parameters).   

\begin{table*}[!ht]
\renewcommand{\arraystretch}{1.2}
\caption{Recognition accuracy evaluation techniques among the studied face BTP methods. \textbf{Metrics/plots:} False Accept/Reject/Match/Non-Match/Positive Rate (FAR/FRR/FMR/FNMR/FPR), Equal Error Rate (EER), Genuine/True Accept/Match Rate (GAR/TAR/TMR), False Negative/Positive Identification Rate (FNIR/FPIR), True/False Positive/Negative (TP/FP/TN/FN), Area Under Curve (AUC), Identification Rate (IR), Detection and Identification Rate (DIR), Detection Error Trade-off (DET), Receiver Operating Characteristic (ROC), True/Mis-/Failed Identification Rate (TIR/MIR/FIR). Assume ``Accuracy'' = $\frac{TP + TN}{TP + FP + FN + TN}$. \textbf{Comparisons:} ``unprotected'' (original) templates vs. ``protected'' (via other BTP methods) templates. \newline \textit{Comparisons key:} {\color{olive}\ding{52}} = explicit; {\color{blue}$\thickapprox$} = implicit/partial; {\color{red}\ding{56}} = none.} 
\centering
\begin{tabular}{|c|c|c|c|c|c|}
\hline
\multirow{3}{*}{\textbf{Method type}} & \multirow{3}{*}{\textbf{Reference}} & \multicolumn{2}{c|}{\textbf{Evaluated in terms of:}} & \multicolumn{2}{c|}{\textbf{Compared to:}}\\
\cline{3-6}
 & & \multirow{2}{*}{\textbf{Metrics}} & \multirow{2}{*}{\textbf{Plots}} & \textbf{Unprotected} & \textbf{Protected} \\
 & & & & \textbf{templates?} & \textbf{templates?} \\
\hline

\multirow{34}{*}{Non-NN} & \cite{m17} & Accuracy (\%) & {\color{red}\ding{56}} & {\color{olive}\ding{52}} & {\color{olive}\ding{52}} \\

 \cline{2-6}
 
 & \cite{b18} & TAR @ FAR = \{0.01, 0.1, 1\} (\%) & {\color{red}\ding{56}} & {\color{olive}\ding{52}} & {\color{red}\ding{56}} \\
 
 \cline{2-6}
 
 & \cite{b19} & {\color{red}\ding{56}} & DET (FNIR vs. FPIR) (\%) & {\color{blue}$\thickapprox$} & {\color{red}\ding{56}} \\
 
 \cline{2-6}
 
 & \cite{j20} & GAR @ FAR = \{0.01, 0.1, 1\} (\%) & {\color{red}\ding{56}} & {\color{olive}\ding{52}} & {\color{olive}\ding{52}} \\
 
 \cline{2-6}
 
 & \multirow{2}{*}{\cite{d21a, d21}} & Rank-1 recognition rate (\%), & \multirow{2}{*}{DET (FNIR vs. FPIR) (\%)} & \multirow{2}{*}{{\color{blue}$\thickapprox$}} & \multirow{2}{*}{{\color{blue}$\thickapprox$}} \\
 
 & & EER (\%), FNIR @ FPIR = 0.1 (\%) & & & \\

 \cline{2-6}
 
 & \cite{e20, e22} & Rank-1 accuracy (\%) & {\color{red}\ding{56}} & {\color{blue}$\thickapprox$} & {\color{red}\ding{56}} \\
 
 \cline{2-6}
 
 & \multirow{2}{*}{\cite{o21}} & Rank-1 IR (\%) & \multirow{2}{*}{DET (FNIR vs. FPIR) (\%)}  & \multirow{2}{*}{{\color{olive}\ding{52}}} & \multirow{2}{*}{{\color{olive}\ding{52}}} \\
 
 & & FNIR @ FPIR = \{0, 0.1, 1\} (\%) &  &  & \\
 
 \cline{2-6}
 
 & \cite{p17} & {\color{red}\ding{56}} & ROC (FRR vs. FAR, plus EER) (\%) & {\color{red}\ding{56}} & {\color{red}\ding{56}} \\
 
 \cline{2-6}
 
 & \cite{m19} & TAR @ FAR = \{1, 0.1, 0.01, 0.001\} (\%) & ROC (TAR vs. FAR) & {\color{olive}\ding{52}} & {\color{red}\ding{56}} \\
 
 \cline{2-6}
 
 & \cite{ko20, ko21} & {\color{red}\ding{56}} & {\color{red}\ding{56}} & {\color{red}\ding{56}} & {\color{red}\ding{56}} \\
 
 \cline{2-6}
 
 & \cite{g19} & \{TP, FP, TN, FN\} @ FPR = 0.001 & {\color{red}\ding{56}} & {\color{olive}\ding{52}} & {\color{red}\ding{56}} \\
 
 \cline{2-6}
 
 & \cite{r21, r22} & FNMR and FMR (\%) & {\color{red}\ding{56}} & {\color{blue}$\thickapprox$} & {\color{olive}\ding{52}} \\
 
 \cline{2-6}
 
 & \cite{a14} & EER & DET (Miss vs. False Alarm prob.) (\%) & {\color{blue}$\thickapprox$} & {\color{red}\ding{56}} \\
 
 \cline{2-6}
 
 & \cite{a19} & Accuracy (\%) & {\color{red}\ding{56}} & {\color{olive}\ding{52}} & {\color{olive}\ding{52}} \\
 
 \cline{2-6}
 
 & \cite{dm20} & EER (\%) & {\color{red}\ding{56}} & {\color{olive}\ding{52}} & {\color{red}\ding{56}} \\
 
 \cline{2-6}
 
 & \cite{s18} & Recognition rate (\%) & {\color{red}\ding{56}} & {\color{red}\ding{56}} & {\color{red}\ding{56}} \\
 
 \cline{2-6}
 
 & \cite{k20} & Accuracy (\%) & DET (FRR vs. FAR, plus EER) (\%) & {\color{olive}\ding{52}} & {\color{red}\ding{56}} \\
 
 \cline{2-6}
 
 & \multirow{3}{*}{\cite{d19}} & AUC (\%), EER (\%), TPR (\%), & \multirow{3}{*}{{\color{red}\ding{56}}} & \multirow{3}{*}{{\color{olive}\ding{52}}} & \multirow{3}{*}{{\color{olive}\ding{52}}} \\
 
  & & TAR @ FAR = 0.1 (\%),  & & & \\
 
  & & Rank-1 DIR @ FAR = 1 (\%) & & & \\
 
 \cline{2-6}
 
 & \cite{d20} & EER (\%), GAR @ FAR (\%) & {\color{red}\ding{56}} & {\color{red}\ding{56}} & {\color{olive}\ding{52}} \\
 
 \cline{2-6}
 
 & \cite{w20} & FAR and FRR (\%) & {\color{red}\ding{56}} & {\color{red}\ding{56}} & {\color{olive}\ding{52}} \\
 
 \cline{2-6}
 
 & \cite{a20} & EER (\%) & {\color{red}\ding{56}} & {\color{red}\ding{56}} & {\color{red}\ding{56}} \\
 
 \cline{2-6}
 
 & \cite{x20} & EER & ROC (1 - FNMR vs. FMR) & {\color{olive}\ding{52}} & {\color{red}\ding{56}} \\
 
 \cline{2-6}
 
 & \cite{j21} & TAR @ FAR (\%) & {\color{red}\ding{56}} & {\color{olive}\ding{52}} & {\color{olive}\ding{52}} \\
 
 \cline{2-6}
 
 & \multirow{5}{*}{\cite{h21}} & IR (\%) @ Rank = \{1, 10, 50\}, & \multirow{5}{*}{ROC (VR vs. FAR) (\%)} & \multirow{5}{*}{{\color{blue}$\thickapprox$}} & \multirow{5}{*}{{\color{red}\ding{56}}} \\
 
 & & Accuracy (\%), & & & \\
 
 & & Verification Recognition @ FAR = 0.1 (\%), & & & \\
 
 & & Rank-1 DIR @ FAR = 1 (\%), & & & \\
 
 & & Own metrics: TIR, MIR, FIR (\%) & & & \\
 
 \cline{2-6}
 
 & \cite{p19} & Accuracy, FAR, FRR, EER (\%) & {\color{red}\ding{56}} & {\color{olive}\ding{52}} & {\color{olive}\ding{52}} \\
 
 \cline{2-6}
 
 & \cite{kh21, kh22} & 1 - FNMR (TMR) @ FMR = $10^{-3}$ (or 0.1\%) & ROC (1 - FNMR vs. FMR) & {\color{olive}\ding{52}} & {\color{olive}\ding{52}} \\
 
\hline

\multirow{15}{*}{NN-learned} & \cite{p15, p16} & EER (\%), GAR @ FAR = \{0, 1\} (\%) & {\color{red}\ding{56}} & {\color{red}\ding{56}} & {\color{olive}\ding{52}}\\

 \cline{2-6}
 
 & \cite{z19} & EER (\%), GAR @ FAR = 1 (\%) & ROC (GAR vs. FAR) & {\color{red}\ding{56}} & {\color{olive}\ding{52}} \\
 
 \cline{2-6}
 
 & \cite{j18} & EER (\%), GAR @ FAR = \{0, 0.01, 0.1\} (\%) & ROC (GAR vs. FAR) & {\color{red}\ding{56}} & {\color{olive}\ding{52}} \\
 
 \cline{2-6}
 
 & \cite{jc19} & EER (\%), GAR @ FAR = 0 (\%) & {\color{red}\ding{56}} & {\color{red}\ding{56}} & {\color{olive}\ding{52}} \\
 
 \cline{2-6}
 
 & \cite{j19} & EER (\%), GAR @ FAR (\%) & {\color{red}\ding{56}} & {\color{red}\ding{56}} & {\color{olive}\ding{52}} \\
 
 \cline{2-6}
 
 & \cite{r18} & FAR and FRR (\%) & {\color{red}\ding{56}} & {\color{red}\ding{56}} & {\color{red}\ding{56}} \\
 
 \cline{2-6}
 
 & \cite{z20} & EER & {\color{red}\ding{56}} & {\color{red}\ding{56}} & {\color{olive}\ding{52}} \\
 
 \cline{2-6}
 
 & \cite{t19} & EER (\%), GAR @ FAR = 0.01 (\%) & ROC (GAR vs. FAR) (\%) & {\color{red}\ding{56}} & {\color{olive}\ding{52}} \\
 
 \cline{2-6}
 
 & \cite{s21} & Accuracy (\%), GAR @ FAR = 0 (\%) & ROC (TAR vs. FAR) & {\color{olive}\ding{52}} & {\color{olive}\ding{52}}  \\
 
 \cline{2-6}
 
 & \multirow{2}{*}{\cite{c19}} & EER (\%), GAR @ FAR = 0.1 (\%), & \multirow{2}{*}{{\color{red}\ding{56}}} & \multirow{2}{*}{{\color{red}\ding{56}}} & \multirow{2}{*}{{\color{olive}\ding{52}}} \\
 
 & & Maximum Average Precision & & & \\
 
 \cline{2-6}
 
 & \cite{m21} & GAR @ FAR = 0.1 (\%) & {\color{red}\ding{56}} & {\color{olive}\ding{52}} & {\color{olive}\ding{52}} \\
 
 \cline{2-6}
 
 & \cite{p21} & EER (\%), FNMR @ FMR = 0.1 (\%) & DET (FNMR vs. FMR) & {\color{olive}\ding{52}} & {\color{olive}\ding{52}} \\
 
 \cline{2-6}
 
 & \cite{c20} & EER (\%) & {\color{red}\ding{56}} & {\color{olive}\ding{52}} & {\color{olive}\ding{52}}  \\
 
 \cline{2-6}
 
 & \cite{l21} & EER (\%) & ROC (TPR vs. FPR) & {\color{olive}\ding{52}} & {\color{olive}\ding{52}} \\
 
\hline
\end{tabular}
\end{table*}

From Table VIII, it is evident that most existing face BTP methods have been proposed for \textit{verification} (as opposed to \textit{identification}) systems, since the majority of employed metrics/plots are verification-system-based. The two most popular evaluation metrics appear to be: (i) the EER, which is the point at which the FAR/FMR and FRR/FNMR are equal, and (ii) the GAR/TAR/TMR reported at the match threshold corresponding to a particular FAR/FMR (i.e., GAR/TAR/TMR @ FAR/FMR), with an FAR/FMR of 0.1\% being the most common threshold choice. The most frequently used plots, which portray the recognition accuracy for a range of different thresholds, include the standard DET and ROC curves, albeit with some inconsistencies in the axis labels (e.g., for ROC, sometimes the \textit{y}-axis was FRR and sometimes it was GAR = 1 - FRR). For the few BTP methods that were evaluated in an \textit{identification} scenario, some form of Rank-1 Identification/Recognition Rate/Accuracy was the most commonly adopted metric, and DET plots with FNIR on the \textit{y}-axis and FPIR on the \textit{x}-axis were the most frequently used plots. 

Table VIII also indicates which BTP methods' evaluated recognition accuracy was compared to the accuracy resulting from using: (i) \textit{unprotected} templates, and (ii) face templates protected using \textit{other} BTP methods. Note that an \textit{explicit} comparison (indicated by {\color{olive}\ding{52}}) means that the evaluation results of the ``other'' systems (i.e., those using unprotected templates or templates protected by other BTP methods) were deliberately/directly compared (e.g., in a table, on the same plot, or conceptually). On the other hand, an \textit{implicit/partial} comparison ({\color{blue}$\thickapprox$}) means that either there was some analysis/mention of the recognition accuracy of the ``other'' system(s) but there was no explicit comparison, or the comparison was presented for only a part of the protected system (e.g., in a hybrid BTP method with multiple stages of protection). 

Comparing the accuracy of a face recognition system using \textit{protected} templates against the same system using \textit{unprotected} templates would provide a good indication of how the adopted BTP method affects the baseline (without BTP) system's recognition accuracy; indeed, such a comparison is traditionally implied in the evaluation of a BTP method's ``recognition accuracy'' criterion. From Table VIII, we can see that this comparison to \textit{unprotected} face templates was conducted for the majority of \textit{Non-NN} face BTP methods but for only a few \textit{NN-learned} BTP methods. This may be attributed to the fact that Non-NN BTP methods are essentially \textit{add-ons} to an existing face recognition system, since they are usually incorporated into only one stage (most commonly after the feature extractor). This means that the BTP part of the protected system can simply be excluded when evaluating the recognition accuracy of the unprotected system. NN-learned BTP methods, on the other hand, tend to be more intertwined within the face recognition system, particularly those that are trained end-to-end. Consequently, it may be difficult to remove the BTP part of the NN to emulate the unprotected system and thereby evaluate its recognition accuracy. 

Of the NN-learned BTP works that provided an explicit recognition accuracy comparison to \textit{unprotected} templates, \cite{s21, p21, c20, l21} performed this evaluation by directly accessing these templates. In particular, the unprotected face templates came from the individual (pre-fusion) NNs in \cite{s21}, the NN trained using the original Triplet Loss instead of the proposed Secure Triplet Loss formulation in \cite{p21}, and straight from the pre-trained DNN that was used as the feature extractor preceding the BTP NN in \cite{c20, l21}. Alternatively, for the end-to-end trained BTP NN in \cite{m21}, the comparison was based on unprotected templates that were extracted using \textit{other} ``state-of-the-art'' feature extractors. It seems that a more direct comparison could have been possible by extracting the unprotected face templates from the \textit{same} NN, using only the Feature Extraction part of the Randomized CNN; however, the adopted comparison approach is interesting, because it seems to imply that the Randomized CNN is considered as a \textit{whole} biometric system that is not to be split up into unprotected/protected sub-systems. In this case, it appears reasonable to compare the recognition accuracy of the Randomized CNN to that of other, unprotected systems. To ensure fairness, the comparisons in \cite{m21} were based on the same face dataset, using the same comparison function and verification protocol.

Considering comparisons to face templates protected using \textit{other} BTP methods in Table VIII, we see that the majority of \textit{NN-learned} BTP works presented such a comparison, in contrast to fewer than half of \textit{Non-NN} BTP works. This is probably because many of the NN-learned BTP methods build upon each other, so it makes sense to compare the recognition accuracy across these methods. For example, \cite{z19, j18, jc19, j19, t19, c19, m21} all compared the accuracy of their proposed BTP methods to one or both of \cite{p16, j18}, either because they are improved versions of \cite{p16, j18} or because these methods are among the first NN-learned BTP approaches so they are unanimously considered as state-of-the-art. A similar progression of improvements is not as evident among the Non-NN BTP methods, however, so a comparison of the recognition accuracy to that of other face BTP methods was more likely to be avoided. When such a comparison \textit{was} provided, it tended to be against traditional methods that the authors deemed similar to their proposed BTP method (e.g., HE, BioHashing, Fuzzy Vault, etc.), or on those methods for which the recognition accuracy had been reported on the same face dataset. 

For the majority of both Non-NN and NN-learned works that presented a comparison to other BTP methods, the comparisons were based on \textit{reported} results generated on the \textit{same} face dataset. The use of the same dataset is important to ensure fairness in the comparison, as is the use of the same \textit{evaluation protocol} (e.g., same train/test images, same reference/probe samples, etc.). Since the comparisons noted in Table VIII were mostly based on \textit{reported} results (as opposed to being generated using shared/own implementations of the other BTP methods), it is not always clear whether exactly the same protocol was applied when evaluating the recognition accuracy. For example, \cite{d20, j18, j19, z19, jc19, t19} all compared the recognition accuracy of their BTP methods to that of \cite{p16} using the CMU PIE face dataset. All publications mention that the same 5 ``poses'' (and all illumination variations) were selected for the protocol, but that the train/test images (within those pose categories) were randomly chosen. In this case, although the protocol is \textit{close enough} to that in \cite{p16}, it does not appear to be \textit{the same}, since the train/test splits were \textit{random}. Most of these works attempted to account for this by presenting the \textit{average} (and sometimes $\pm$ standard deviation) of their EER and GAR @ FAR results across 5 random train/test splits. So, the comparison seems \textit{fair enough}; however, to ensure 100\% fairness in general, we must make sure to use the same dataset, along with exactly the same comparison protocol, when computing the recognition accuracy. 

It should be noted that a few of the studied works (\cite{r21, r22, d21a, d21, m21, s21}) reported some recognition accuracy comparisons using results generated on datasets that were \textit{different} to those used to evaluate their proposed BTP methods. While such comparisons may have been included for completeness' sake, and the authors explicitly noted the use of different datasets, in general the usefulness of presenting inter-dataset comparisons is doubtful. This is because, unless the recognition accuracy for the methods being compared has been evaluated under the \textit{same conditions}, the comparison cannot be considered fair and can, in fact, often be misleading. So, we would recommend that such comparisons be avoided. A suitable alternative could be to provide some insights into the comparison from a \textit{conceptual} point of view (e.g., how the proposed BTP method may be \textit{expected} to perform compared to another method), as presented in \cite{kh21, kh22}. This approach could be adopted until it is possible to provide a fair \textit{numerical} comparison (or it could be used to complement the numerical comparison). Provided that this analysis is performed honestly, it would probably be more meaningful than an unfair numerical comparison. 

On a final note, to make comparisons between different BTP methods easier and more useful, we encourage the BTP community to strive for the \textit{reproducibility} of both their BTP method implementations and their evaluation procedures. This way, firstly, we could ensure that comparisons on the same dataset are based on exactly the same evaluation protocol. Secondly, we would have more freedom in the choice of dataset used for the comparison, instead of feeling compelled to either constrain our evaluations to only those datasets on which the results of other BTP methods have been reported or to present inter-dataset comparisons. This is because, if we had access to open-source implementations of other researchers' BTP methods, we could evaluate those methods on whatever dataset(s) we deem to be most representative of our proposed BTP methods' intended operational scenario(s). This would help us provide fairer and more meaningful insights into how our BTP method compares to other BTP methods \textit{in the most appropriate evaluation context}. The reproducibility of the existing face BTP methods was considered in Section III.A.   

\subsection{Irreversibility}

A BTP method is considered to satisfy the ``irreversibility'' criterion if it is impossible (or computationally infeasible) to recover the original biometric template (features or image) from the protected template. This section investigates which face BTP methods from Section II were evaluated for their ability to satisfy this criterion and what techniques were used to conduct the evaluation. 

Table IX summarises our findings, in terms of indicating whether  irreversibility for each face BTP method was evaluated \textit{theoretically} or \textit{empirically} \footnote{As per the ISO/IEC standard 30136 on \textit{Performance testing of biometric template protection schemes} (\url{https://bit.ly/398Az9k}), which broadly suggests either a theoretical or empirical evaluation of irreversibility.} (if at all). In general, theoretical evaluations include proofs or quantified estimates of irreversibility based on certain mathematical, cryptographic, or information-theoretic assumptions; however, no experiments are performed to simulate any inversion attacks in practice. Empirical evaluations, on the other hand, attempt to prove or quantify irreversibility in terms of specific attacks, which are simulated through practical experiments. 

\begin{table*}[!ht]
\renewcommand{\arraystretch}{1.2}
\caption{Irreversibility evaluation approaches among the studied face BTP methods. \textbf{Theoretical:} Proofs/estimates of irreversibility without attack simulations. \textbf{Empirical:} Proofs/estimates of irreversibility via experiments simulating specific attacks. \newline \textit{Evaluations:} {\color{olive}\ding{52}} = Some (numerical/conceptual); {\color{red}\ding{56}} = None. \textit{Comparisons:} {\color{olive}\ding{52}} = explicit; {\color{blue}$\thickapprox$} = implicit/partial; {\color{red}\ding{56}} = none.} 
\centering
\begin{tabular}{|c|c|c|c|c|}
\hline
\multirow{2}{*}{\textbf{Method type}} & \multirow{2}{*}{\textbf{Reference}} & \multicolumn{2}{c|}{\textbf{Irreversibility evaluated:}} & \multirow{2}{*}{\textbf{Compared to other BTP methods?}} \\
\cline{3-4}
 & & \textbf{Theoretically} & \textbf{Empirically} & \\
\hline

\multirow{26}{*}{Non-NN} & \cite{m17} & {\color{olive}\ding{52}} &  {\color{red}\ding{56}} & {\color{red}\ding{56}} \\

 \cline{2-5}
 
 & \cite{b18} & {\color{olive}\ding{52}} & {\color{red}\ding{56}} & {\color{red}\ding{56}} \\
 
 \cline{2-5}
 
 & \cite{b19} & {\color{olive}\ding{52}} & {\color{red}\ding{56}} & {\color{red}\ding{56}} \\
 
 \cline{2-5}
 
 & \cite{j20} & {\color{olive}\ding{52}} & {\color{red}\ding{56}} & {\color{red}\ding{56}} \\
 
 \cline{2-5}
 
 & \cite{d21a, d21} & {\color{olive}\ding{52}} & {\color{red}\ding{56}} & {\color{red}\ding{56}} \\

 \cline{2-5}
 
 & \cite{e20, e22} & {\color{olive}\ding{52}} & {\color{red}\ding{56}} & {\color{red}\ding{56}} \\

 \cline{2-5}
 
 & \cite{o21} & {\color{olive}\ding{52}} & {\color{red}\ding{56}} & {\color{red}\ding{56}} \\

 \cline{2-5}
 
 & \cite{p17} & {\color{olive}\ding{52}} & {\color{red}\ding{56}} & {\color{red}\ding{56}} \\
 
 \cline{2-5}
 
 & \cite{m19} & {\color{red}\ding{56}} & {\color{red}\ding{56}} & {\color{red}\ding{56}} \\
 
 \cline{2-5}
 
 & \cite{ko20, ko21} & {\color{red}\ding{56}} & {\color{olive}\ding{52}} & {\color{red}\ding{56}} \\
 
 \cline{2-5}
 
 & \cite{g19} & {\color{olive}\ding{52}} & {\color{red}\ding{56}} & {\color{red}\ding{56}} \\
 
 \cline{2-5}
 
 & \cite{r21, r22} & {\color{red}\ding{56}} & {\color{olive}\ding{52}} & {\color{olive}\ding{52}} \\
 
 \cline{2-5}
 
 & \cite{a14} & {\color{red}\ding{56}} & {\color{red}\ding{56}} & {\color{red}\ding{56}} \\
 
 \cline{2-5}
 
 & \cite{a19} & {\color{red}\ding{56}} & {\color{olive}\ding{52}} & {\color{red}\ding{56}} \\
 
 \cline{2-5}
 
 & \cite{dm20} & {\color{red}\ding{56}} & {\color{red}\ding{56}} & {\color{red}\ding{56}} \\
 
 \cline{2-5}
 
 & \cite{s18} & {\color{olive}\ding{52}} & {\color{red}\ding{56}} & {\color{red}\ding{56}} \\
 
 \cline{2-5}
 
 & \cite{k20} & {\color{red}\ding{56}} & {\color{olive}\ding{52}} & {\color{red}\ding{56}} \\
 
 \cline{2-5}
 
 & \cite{d19} & {\color{olive}\ding{52}} & {\color{red}\ding{56}} & {\color{olive}\ding{52}} \\
 
 \cline{2-5}
 
 & \cite{d20} & {\color{olive}\ding{52}} & {\color{red}\ding{56}} & {\color{red}\ding{56}} \\
 
 \cline{2-5}
 
 & \cite{w20} & {\color{olive}\ding{52}} & {\color{red}\ding{56}} & {\color{red}\ding{56}} \\
 
 \cline{2-5}
 
 & \cite{a20} & {\color{red}\ding{56}} & {\color{red}\ding{56}} & {\color{red}\ding{56}} \\
 
 \cline{2-5}
 
 & \cite{x20} & {\color{red}\ding{56}} & {\color{red}\ding{56}} & {\color{red}\ding{56}} \\
 
 \cline{2-5}
 
 & \cite{j21} & {\color{olive}\ding{52}} & {\color{red}\ding{56}} & {\color{red}\ding{56}} \\
 
 \cline{2-5}
 
 & \cite{h21} & {\color{olive}\ding{52}} & {\color{red}\ding{56}} & {\color{red}\ding{56}} \\
 
 \cline{2-5}
 
 & \cite{p19} & {\color{olive}\ding{52}} & {\color{red}\ding{56}} & {\color{red}\ding{56}} \\
 
 \cline{2-5}
 
 & \cite{kh21, kh22} & {\color{olive}\ding{52}} & {\color{olive}\ding{52}} & {\color{red}\ding{56}} \\
 
\hline

\multirow{14}{*}{NN-learned} & \cite{p15, p16} & {\color{olive}\ding{52}} & {\color{olive}\ding{52}} & {\color{olive}\ding{52}} \\

 \cline{2-5}
 
 & \cite{z19} & {\color{olive}\ding{52}} & {\color{olive}\ding{52}} & {\color{olive}\ding{52}} \\
 
 \cline{2-5}
 
 & \cite{j18} & {\color{olive}\ding{52}} & {\color{olive}\ding{52}} & {\color{blue}$\thickapprox$} \\
 
 \cline{2-5}
 
 & \cite{jc19} & {\color{olive}\ding{52}} & {\color{olive}\ding{52}} & {\color{blue}$\thickapprox$} \\
 
 \cline{2-5}
 
 & \cite{j19} & {\color{olive}\ding{52}} & {\color{red}\ding{56}} & {\color{blue}$\thickapprox$} \\
 
 \cline{2-5}
 
 & \cite{r18} & {\color{red}\ding{56}} & {\color{red}\ding{56}} & {\color{red}\ding{56}} \\
 
 \cline{2-5}
 
 & \cite{z20} & {\color{olive}\ding{52}} & {\color{red}\ding{56}} & {\color{blue}$\thickapprox$} \\
 
 \cline{2-5}
 
 & \cite{t19} & {\color{olive}\ding{52}} & {\color{olive}\ding{52}} & {\color{olive}\ding{52}} \\
 
 \cline{2-5}
 
 & \cite{s21} & {\color{red}\ding{56}} & {\color{olive}\ding{52}} & {\color{red}\ding{56}} \\
 
 \cline{2-5}
 
 & \cite{c19} & {\color{olive}\ding{52}} & {\color{red}\ding{56}} & {\color{olive}\ding{52}} \\
 
 \cline{2-5}
 
 & \cite{m21} & {\color{olive}\ding{52}} & {\color{red}\ding{56}} & {\color{red}\ding{56}} \\
 
 \cline{2-5}
 
 & \cite{p21} & {\color{olive}\ding{52}} & {\color{red}\ding{56}} & {\color{red}\ding{56}} \\
 
 \cline{2-5}
 
 & \cite{c20} & {\color{olive}\ding{52}} & {\color{red}\ding{56}} & {\color{red}\ding{56}} \\
 
 \cline{2-5}
 
 & \cite{l21} & {\color{olive}\ding{52}} & {\color{red}\ding{56}} & {\color{red}\ding{56}} \\
 
\hline
\end{tabular}
\end{table*} 

From Table IX, it is clear that the irreversibility of most of the studied face BTP methods was evaluated \textit{theoretically}. The exact evaluation strategy depended on the nature of the BTP method, so it would be tedious to present a detailed examination of each technique here. Instead, we will discuss the most widely adopted theoretical evaluation methods.

For the methods proposed in \cite{m17, b18, b19, j20, d21a, d21, e20, e22, o21}, which are based on homomorphic encryption (HE), their irreversibility was attributed\footnote{HE-protected templates in \cite{o21} are indexed by binary codes generated from the original templates, so it is suggested that the codes be protected by cryptographic hashing. Code irreversibility is assumed based on the hashing.} to: (i) the strength of the HE algorithm used to secure the face templates, and (ii) the secrecy of the corresponding decryption key. In this case, if the decryption key is leaked to an attacker, then the protected (encrypted) face templates become completely reversible. Otherwise, the ``degree'' of irreversibility was typically quantified in terms of the known security of the employed HE scheme; for example, the upper-bound would be the length of the decryption key in bits, $n$, which implies that at most $2^n$ guesses would be required to guess the key using a brute-force approach. 

The irreversibility of the face BTP methods in \cite{p17, m19, d20, j21, p15, p16, z19, j18, jc19, j19, t19} was largely attributed to the use of a one-way cryptographic hash function. So, it was generally assumed that the only way to ``invert'' the hash was to attempt a brute-force attack, whereby all possible inputs to the hashing function are tried until the target hash is obtained. For an $n$-bit input, this amounts to an upper-bound of $2^n$ guesses. Note that this is the same complexity as that for brute-force guessing the decryption key in HE schemes (discussed earlier), except that this time the attacker is trying to guess the \textit{input} to the hashing function instead of the hash itself (since the hash is not secret). The cryptographic hash function in \cite{p17, t19} was used to directly secure the (binarised) face features, meaning that the input space for the brute-force attack would be the binary face feature domain. In this case, $2^n$ would probably be an over-estimation of the brute-force attack complexity, since the binarised face features would not be uniformly distributed (i.e., there is unlikely to be $2^n$ equally probable binary feature vectors to guess from). On the other hand, the BTP methods in \cite{m19, j21, p15, p16, z19, j18, jc19, j19, t19} hashed a \textit{random codeword} that was either: (i) bound with the face feature vector \cite{m19, j21}, or (ii) designated as the pre-hash protected face template, to which the original feature vector was then mapped \cite{p15, p16, z19, j18, jc19, j19}. In this case, the inputs in a brute-force attack would come from the codeword domain; so, provided that the codewords are generated from a uniform distribution, $2^n$ is a fair approximation of the brute-force attack complexity.  

Variations of brute-force or guessing attack complexity were also used to theoretically estimate the irreversibility of face BTP methods that do not rely (primarily) on cryptographic hash functions, including \cite{w20, h21, c19, c20, l21}. In general, however, basing an irreversibility evaluation on the assumption of a brute-force attack would most probably result in an over-estimation of the irreversibility in practice. This is because there will often exist other, easier attacks, which would allow an adversary to invert the protected template faster than using a brute-force approach. So, one way to ``tone down'' the brute-force attack complexity estimate could be to approximate the \textit{entropy} of the input space (instead of assuming a uniform distribution) and/or the reduction in the attacker's \textit{uncertainty} about the input space (which would help them make more informed guesses). The latter is usually estimated in terms of the amount of information leaked by the protected template or certain auxiliary data, which is the ``mutual information'' irreversibility evaluation method mentioned in the ISO/IEC standard 30136 on \textit{Performance testing of biometric template protection schemes}. Variations of the entropy and mutual information irreversibility evaluation approaches were adopted in \cite{p17, d19, z19, m21, p21}.

Alternatively, we could turn to \textit{empirical} irreversibility evaluation strategies, which focus on practical attack simulations. Ideally, such evaluations should demonstrate the existence (or lack thereof) of attacks easier than brute-force, which could be used to invert a protected template in practice. For example, of the aforementioned ``binding'' BTP methods that cryptographically hash a random codeword, guessing the codeword would be instrumental in ``unbinding'' the protected template and thus recovering the original template. Instead of brute-force guessing the codeword, however, the Fuzzy Commitment scheme in \cite{m19} may be invertible using the approach in \cite{ko20, ko21}, while \cite{j21} is stated to be invertible if the user-specific linear transformation matrix and the face template distribution are known. For the BTP methods that hash a \textit{pre-defined} codeword to represent the protected template (\cite{p15, p16, z19, j18, jc19, j19}), guessing this codeword from its hash would, on its own, provide no information about the original face template since the two are unrelated. So, although the brute-force attack complexity seems to be a more appropriate measure of irreversibility for these types of BTP methods, it is still unlikely that a brute-force attack would be the only avenue for a determined attacker. For example, we may imagine that a fully-informed attacker with access to the trained neural network and all its parameters, along with a face dataset that is representative of the enrolled subjects, could exploit this information to determine how the neural network learns a mapping between a face template and a randomly chosen codeword. Note that, as discussed in Section II.B, a neural network trained to map a face template to a pre-defined random codeword is unlikely to be generalisable to unseen (unenrolled) subjects. This implies that access to the trained network without access to the un-hashed codewords may not directly reveal a potential inverse mapping. Nevertheless, this type of analysis may allow the attacker to extract certain information about the mapping function in certain layers of the neural network, which could narrow down the search space for the un-hashed codeword and its corresponding face template. Although this evaluation may not be trivial, it should still be considered as a worst-case scenario, since such avenues are more likely to prove fruitful for a fully-informed attacker (compared to attempting a brute-force attack on the cryptographic hash itself). 

It should be noted that \cite{p15, p16, z19, j18} presented an alternative evaluation of irreversibility, besides a brute-force attack on the cryptographic hash. In particular, they simulated a dictionary attack, where an adversary with access to the trained neural network attempts a brute-force attack in the image domain to exploit the system's FAR, i.e., the attacker presents face images to the protected system in the hope of finding a match to an enrolled identity. The images used to simulate the attack, however, came from a different dataset than those used to generate the enrolled protected templates. This means that the ``attack'' images cannot be considered representative of the face images corresponding to the enrolled subjects, so it is no surprise that the dictionary attack failed. A similar evaluation was presented in \cite{s21}. In \cite{jc19, j19}, a dictionary attack was deemed unnecessary due to the zero FAR achieved in the recognition accuracy evaluation. This low error rate, however, may be attributed in large part to the use of user-specific transformations (random projection in \cite{jc19} and random perturbations in \cite{j19}), so an FAR of 0 may not be observed if the transformation parameters are leaked to the attacker.

Besides a dictionary attack, other examples of empirical evaluations that focus on exploiting the biometric system's FAR, were presented in \cite{r21, r22, k20}. In \cite{r21, r22}, the irreversibility of the proposed BTP method was estimated (in bits) as: $\texttt{False Accept Security (FAS)} = l \times \log(0.5) / \log(1 - FMR)$ (where $l$ = the average number of operations for a non-mated verification attempt). The irreversibility of the BTP method in \cite{k20} was also evaluated in terms of an empirical system attack, except that this time there was no guessing of the inputs. Instead, the protected (warped) face images in \cite{k20} were directly compared to their unprotected counterparts, demonstrating that insufficient warping results in a high match accuracy.

Unquantified empirical irreversibility evaluations were presented in \cite{a19, s18}, which showed visual examples of face images that were ``encrypted'' using the correct user-specific parameters and ``decrypted'' to a noise-like output using the wrong parameters. This suggests that the irreversibility of these BTP methods depends on keeping the user-specific transformation parameters secret (like for HE or traditional encryption methods), but it does not give us an indication of the irreversibility if the parameters are leaked to an attacker.   

Finally, only \cite{ko20, ko21, kh21, kh22} presented an empirical irreversibility analysis in terms of an actual \textit{inversion attack}. This means that, instead of simply guessing inputs that might generate the protected output, \cite{ko20, ko21, kh21, kh22} actually attempted to \textit{invert the protection mechanism} to recover the unprotected face template from the protected template. As mentioned earlier, \cite{ko20, ko21} demonstrated that the Fuzzy Commitment (FC) scheme protecting the face template can be broken. This was performed by first using a random set of face images to guess a binary template that matches the one used in the FC binding, then using the guessed binary template to reconstruct the corresponding real-valued face feature vector, and finally using the recovered feature vector to reconstruct a close approximation of the corresponding face image. The reconstruction steps were achieved using neural networks trained for the inversion. Alternatively, the BTP method in \cite{kh21, kh22}, which was defined via a set of user-specific multivariate polynomials, was evaluated by attempting to invert the polynomial mappings using a numerical solver (assuming a fully informed attacker, with access to all parameters and knowledge of a representative face feature distribution). The recovered real-valued feature vector was compared to the original feature vector, which was assumed to be enrolled in an unprotected face recognition system, based on the system's match threshold. The empirical irreversibility of the BTP method was quantified in terms of the $\texttt{inversion success rate} = \texttt{solution rate} \times \texttt{match rate}$. A similar quantification, based on the closeness between the recovered and original face templates, was also performed in \cite{ko20, ko21}. For both methods, the face data used to obtain the required ``guesses'' (i.e., the binary templates in \cite{ko20, ko21} and the initial guesses for the numerical solver in \cite{kh21, kh22}) came from the same dataset as the enrolled subjects (thereby fairly representing the underlying face feature distributions), except that a different set of subjects was used for this purpose.

The last observation from Table IX is that hardly any of the studied face BTP works presented an explicit comparison of the irreversibility of their method to that of other face BTP methods. Of those that \textit{did}, \cite{r21, r22} compared the irreversibility in terms of the proposed FAS measure to the irreversibility estimates reported for other face-based Fuzzy Vault BTP methods. Most of these other methods, however, were evaluated in terms of brute-force attack complexity, and the evaluations were performed on different datasets using different feature extractors. So, the presented irreversibility comparison does not seem fair. In \cite{d19}, the entropy-based measure of irreversibility was compared to that of a couple of other BTP methods. Even though the comparison seems reasonable in that the same metric was employed, it is not explained how the parameters of these different methods were selected to ensure that the resulting irreversibility strengths were fairly compared. The irreversibility of the methods in \cite{p15, p16, z19, t19} was compared to that of other BTP methods based on ``bits of security'', which was represented by the length of their binary codes. This is because, as discussed earlier, the irreversibility of these methods was evaluated in terms of brute-force attack complexity, which is equivalent to $2^n$ for $n$-bit binary codes. One issue with this comparison (apart from the fact that brute-force attack complexity is likely to be an over-estimate of the true irreversibility) is that it is not always clear: (i) how the \textit{equivalent} code length parameter was established for all the other methods, and (ii) whether the value of the chosen code length parameter is a valid measure of the irreversibility of all these other methods.

Finally, Table IX indicates that \cite{j18, jc19, j19, z20} presented implicit or partial comparisons of the irreversibility of their proposed BTP methods to that of other face BTP methods in the literature. More specifically, similarly to \cite{p15, p16, z19, t19}, the works in \cite{j18, jc19, j19} also provided tables with the ``code length'' of the binary codes resulting from different BTP methods; however, they did not explicitly relate these parameter values to a comparison of the methods' irreversibility (or security). On the other hand, \cite{z20} presented a subjective (in terms of: High, Medium, Low) comparison of the irreversibility of their method to that of other data protection methods, but the chosen methods were purely cryptographic hashing and encryption functions as opposed to (face) BTP algorithms.

Although not noted in Table IX, we also looked at which face BTP methods evaluated irreversibility in terms of a Record Multiplicity Attack (ARM). This refers to the scenario where an attacker is assumed to have access to multiple protected templates from the same subject's face, and they combine this information to try to invert the protected templates. Of the face BTP methods studied in this paper, ARM was evaluated only for \cite{c20, m21, h21, p19, kh21, kh22}. A theoretical ARM analysis was presented\footnote{For \cite{c20, h21}, we are referred to \cite{h18}, on which \cite{c20, h21} are based.} in \cite{c20, m21, h21}, where it was concluded that this type of attack cannot reduce the security of the protected system; however, in all cases, the assumption was that certain user-specific parameters are not leaked to the attacker. ARM was also evaluated\footnote{We are referred to \cite{s13}, where \cite{p19} was applied to iris biometrics.} from a theoretical angle in \cite{p19}, where various scenarios, with different information assumed to have been leaked to the attacker, were considered. The analysis was based on showing that the attack would require solving underdetermined systems of equations, which have no unique solution; however, there was no practical attempt at solving the systems using a numerical solver, unlike in \cite{kh21, kh22}. In \cite{kh21, kh22}, an \textit{empirical} ARM evaluation was presented for the scenario of a fully-informed attacker with access to all system parameters. The evaluation adopted a numerical solver to solve a system of non-linear equations, which represented the (inverse) mapping from each stolen protected template to the original template. As expected, the protected system was shown to be somewhat susceptible to ARM in the worst-case scenario of a fully-informed attacker; however, it is doubtful that \textit{any} BTP method would be able to perfectly resist this type of attack in practice if the attacker had access to everything. 

Overall, our investigation into irreversibility evaluation techniques among the existing face BTP methods revealed that several interesting strategies have been employed, depending on the nature of the BTP algorithm. Many of these techniques, however, may result in inadequate or misleading representations of the irreversibility in practice, especially in the worst-case scenario of a fully-informed attacker (which is the most difficult threat model outlined in the ISO/IEC standard 30136 on \textit{Performance testing of biometric template protection schemes}). This is usually because the protection method is assumed to rely on the secrecy of certain parameters, or because the degree of irreversibility is over-estimated as a result of limited investigation into attacks that are easier/smarter than brute-force. For example, for NN-learned BTP methods, it is conceivable that an attacker with access to the trained BTP neural network (i.e., its architecture and all learned parameters) could use this knowledge to extract certain information about the learned protection algorithm in different layers of the network, which could in turn be used to uncover hints about certain characteristics of the unprotected face templates. This type of analysis is currently lacking in the face BTP literature. 

More generally, based on the summary in Table IX, we conclude that there is a noticeable shortage of \textit{empirical} irreversibility evaluations. Empirical evaluations encourage us to consider the irreversibility of our BTP methods from the point of view of a determined attacker, thereby providing more insight into the strength of these methods in practice (albeit usually in a worst-case scenario). Such practically-oriented types of evaluations may be expected to become increasingly important in the future, considering the growing interest in protecting face templates in real-world face recognition systems. Furthermore, empirical evaluations may make it easier to compare the irreversibility of different BTP methods in more realistic terms. We would, therefore, like to encourage the BTP community to invest more effort towards developing irreversibility evaluation strategies that will: (i) better reflect the expected irreversibility of our BTP methods in practice, and (ii) allow for more meaningful comparisons of the irreversibility of different BTP methods.

\subsection{Renewability/Unlinkability}

A BTP method is considered to satisfy the ``renewability/unlinkability'' criterion if it can generate multiple \textit{diverse} protected templates from the same original template, such that they \textit{cannot be linked} to the same identity. This would allow for compromised templates to be \textit{cancelled/revoked} and \textit{renewed}, as well as enabling the same biometric characteristic to be used in different applications without the risk of cross-matching the enrolled protected templates. 

Note that the ISO/IEC standard 30136 on \textit{Performance testing of biometric template protection schemes} defines ``diversity'' as the expected value of the number of independent protected templates that can be extracted from the same biometric identity, which directly ensures ``renewability'' and ``revocability'' (or ``cancellability''). ``Unlinkability'' is then defined separately as the property that two protected templates cannot be linked to each other or to the subject(s) from which they were derived. In the face BTP literature, however, these terms often seem to be used interchangeably, which is understandable considering the inter-dependence of the underlying concepts. So, for the purposes of this literature survey, we consider all the aforementioned terms to define essentially the same property or criterion of a face BTP method, which we refer to as ``renewability/unlinkability'', i.e., can the BTP method ensure the \textit{renewability} of protected templates by being able to create multiple \textit{unlinkable} protected templates from the same face identity? This section explores whether this criterion was evaluated, and what techniques were used in the analysis, for the face BTP methods from Section II. 

Table X presents the findings of our renewability/unlinkability investigation in terms of indicating whether this criterion was evaluated via \textit{verbal implication}, \textit{theoretical proof}, or \textit{empirical validation} (if at all) for each face BTP method. Verbal implications include statements implying that the criterion is satisfied (either directly or indirectly), but without presenting any theoretical or empirical proof. Theoretical proofs include mathematical reasoning for why the criterion is believed to be fulfilled, or theoretical estimations of the diversity. Finally, empirical validations include experimental attempts to prove and/or quantify the extent to which this criterion is satisfied in a practical setting.

\begin{table*}[!ht]
\renewcommand{\arraystretch}{1.2}
\caption{Renewability/unlinkability evaluation approaches among the studied face BTP methods. \textbf{Verbal implication:} Statement(s) implying that the criterion is satisfied (no analysis). \textbf{Theoretical proof:} Mathematical reasoning or theoretical estimate(s) proving that the criterion is satisfied. \textbf{Empirical validation:} Experiment(s) proving or quantifying the extent to which the criterion is satisfied. \textit{Key:} {\color{olive}\ding{52}} = Some evaluation; {\color{red}\ding{56}} = No evaluation.} 
\centering
\begin{tabular}{|c|c|c|c|c|c|}
\hline
\multirow{2}{*}{\textbf{Method type}} & \multirow{2}{*}{\textbf{Reference}} & \multicolumn{3}{c|}{\textbf{Renewability/Unlinkability evaluated via:}} & \multirow{2}{*}{\textbf{Compared to other BTP methods?}} \\
\cline{3-5}
 & & \textbf{Verbal implication} & \textbf{Theoretical proof} & \textbf{Empirical validation} & \\
\hline

\multirow{26}{*}{Non-NN} & \cite{m17} & {\color{red}\ding{56}} & {\color{red}\ding{56}} & {\color{red}\ding{56}} & {\color{red}\ding{56}} \\

 \cline{2-6}
 
 & \cite{b18} & {\color{olive}\ding{52}} & {\color{red}\ding{56}} & {\color{red}\ding{56}} & {\color{red}\ding{56}} \\
 
 \cline{2-6}
 
 & \cite{b19} & {\color{olive}\ding{52}} & {\color{red}\ding{56}} & {\color{red}\ding{56}} & {\color{red}\ding{56}} \\
 
 \cline{2-6}
 
 & \cite{j20} & {\color{olive}\ding{52}} & {\color{red}\ding{56}} & {\color{red}\ding{56}} & {\color{red}\ding{56}} \\
 
 \cline{2-6}
 
 & \cite{d21a, d21} & {\color{olive}\ding{52}} & {\color{red}\ding{56}} & {\color{red}\ding{56}} & {\color{red}\ding{56}} \\

 \cline{2-6}
 
 & \cite{e20, e22} & {\color{olive}\ding{52}} & {\color{red}\ding{56}} & {\color{red}\ding{56}} & {\color{red}\ding{56}} \\

 \cline{2-6}
 
 & \cite{o21} & {\color{olive}\ding{52}} & {\color{red}\ding{56}} & {\color{red}\ding{56}} & {\color{red}\ding{56}} \\

 \cline{2-6}
 
 & \cite{p17} & {\color{red}\ding{56}} & {\color{red}\ding{56}} & {\color{red}\ding{56}} & {\color{red}\ding{56}} \\
 
 \cline{2-6}
 
 & \cite{m19} & {\color{olive}\ding{52}} & {\color{red}\ding{56}} & {\color{red}\ding{56}} & {\color{red}\ding{56}} \\
 
 \cline{2-6}
 
 & \cite{ko20, ko21} & {\color{red}\ding{56}} & {\color{red}\ding{56}} & {\color{olive}\ding{52}} & {\color{red}\ding{56}} \\
 
 \cline{2-6}
 
 & \cite{g19} & {\color{red}\ding{56}} & {\color{olive}\ding{52}} & {\color{red}\ding{56}} & {\color{red}\ding{56}} \\
 
 \cline{2-6}
 
 & \cite{r21, r22} & {\color{red}\ding{56}} & {\color{olive}\ding{52}} & {\color{red}\ding{56}} & {\color{red}\ding{56}} \\
 
 \cline{2-6}
 
 & \cite{a14} & {\color{red}\ding{56}} & {\color{red}\ding{56}} & {\color{red}\ding{56}} & {\color{red}\ding{56}} \\
 
 \cline{2-6}
 
 & \cite{a19} & {\color{red}\ding{56}} & {\color{red}\ding{56}} & {\color{red}\ding{56}} & {\color{red}\ding{56}} \\
 
 \cline{2-6}
 
 & \cite{dm20} & {\color{red}\ding{56}} & {\color{red}\ding{56}} & {\color{red}\ding{56}} & {\color{red}\ding{56}} \\
 
 \cline{2-6}
 
 & \cite{s18} & {\color{red}\ding{56}} & {\color{red}\ding{56}} & {\color{red}\ding{56}} & {\color{red}\ding{56}} \\
 
 \cline{2-6}
 
 & \cite{k20} & {\color{red}\ding{56}} & {\color{red}\ding{56}} & {\color{olive}\ding{52}} & {\color{red}\ding{56}} \\
 
 \cline{2-6}
 
 & \cite{d19} & {\color{red}\ding{56}} & {\color{red}\ding{56}} & {\color{olive}\ding{52}} & {\color{red}\ding{56}} \\
 
 \cline{2-6}
 
 & \cite{d20} & {\color{olive}\ding{52}} & {\color{red}\ding{56}} & {\color{red}\ding{56}} & {\color{red}\ding{56}} \\
 
 \cline{2-6}
 
 & \cite{w20} & {\color{olive}\ding{52}} & {\color{red}\ding{56}} & {\color{red}\ding{56}} & {\color{red}\ding{56}} \\
 
 \cline{2-6}
 
 & \cite{a20} & {\color{red}\ding{56}} & {\color{red}\ding{56}} & {\color{red}\ding{56}} & {\color{red}\ding{56}} \\
 
 \cline{2-6}
 
 & \cite{x20} & {\color{red}\ding{56}} & {\color{red}\ding{56}} & {\color{olive}\ding{52}} & {\color{red}\ding{56}} \\
 
 \cline{2-6}
 
 & \cite{j21} & {\color{red}\ding{56}} & {\color{olive}\ding{52}} & {\color{red}\ding{56}} & {\color{red}\ding{56}} \\
 
 \cline{2-6}
 
 & \cite{h21} & {\color{olive}\ding{52}} & {\color{red}\ding{56}} & {\color{red}\ding{56}} & {\color{red}\ding{56}} \\
 
 \cline{2-6}
 
 & \cite{p19} & {\color{olive}\ding{52}} & {\color{red}\ding{56}} & {\color{red}\ding{56}} & {\color{red}\ding{56}} \\
 
 \cline{2-6}
 
 & \cite{kh21, kh22} & {\color{red}\ding{56}} & {\color{red}\ding{56}} & {\color{olive}\ding{52}} & {\color{olive}\ding{52}} \\
 
\hline

\multirow{14}{*}{NN-learned} & \cite{p15, p16} & {\color{olive}\ding{52}} & {\color{red}\ding{56}} & {\color{red}\ding{56}} & {\color{red}\ding{56}} \\

 \cline{2-6}
 
 & \cite{z19} & {\color{olive}\ding{52}} & {\color{red}\ding{56}} & {\color{red}\ding{56}} & {\color{red}\ding{56}} \\
 
 \cline{2-6}
 
 & \cite{j18} & {\color{olive}\ding{52}} & {\color{red}\ding{56}} & {\color{red}\ding{56}} & {\color{red}\ding{56}} \\
 
 \cline{2-6}
 
 & \cite{jc19} & {\color{olive}\ding{52}} & {\color{red}\ding{56}} & {\color{red}\ding{56}} & {\color{red}\ding{56}} \\
 
 \cline{2-6}
 
 & \cite{j19} & {\color{olive}\ding{52}} & {\color{red}\ding{56}} & {\color{red}\ding{56}} & {\color{red}\ding{56}} \\
 
 \cline{2-6}
 
 & \cite{r18} & {\color{red}\ding{56}} & {\color{red}\ding{56}} & {\color{red}\ding{56}} & {\color{red}\ding{56}} \\
 
 \cline{2-6}
 
 & \cite{z20} & {\color{red}\ding{56}} & {\color{red}\ding{56}} & {\color{red}\ding{56}} & {\color{red}\ding{56}} \\
 
 \cline{2-6}
 
 & \cite{t19} & {\color{red}\ding{56}} & {\color{red}\ding{56}} & {\color{red}\ding{56}} & {\color{red}\ding{56}} \\
 
 \cline{2-6}
 
 & \cite{s21} & {\color{olive}\ding{52}} & {\color{red}\ding{56}} & {\color{red}\ding{56}} & {\color{red}\ding{56}} \\
 
 \cline{2-6}
 
 & \cite{c19} & {\color{olive}\ding{52}} & {\color{red}\ding{56}} & {\color{red}\ding{56}} & {\color{red}\ding{56}} \\
 
 \cline{2-6}
 
 & \cite{m21} & {\color{red}\ding{56}} & {\color{red}\ding{56}} & {\color{olive}\ding{52}} & {\color{red}\ding{56}} \\
 
 \cline{2-6}
 
 & \cite{p21} & {\color{red}\ding{56}} & {\color{red}\ding{56}} & {\color{olive}\ding{52}} & {\color{olive}\ding{52}} \\
 
 \cline{2-6}
 
 & \cite{c20} & {\color{olive}\ding{52}} & {\color{red}\ding{56}} & {\color{red}\ding{56}} & {\color{red}\ding{56}} \\
 
 \cline{2-6}
 
 & \cite{l21} & {\color{red}\ding{56}} & {\color{olive}\ding{52}} & {\color{olive}\ding{52}} & {\color{red}\ding{56}} \\
 
\hline
\end{tabular}
\end{table*} 

There are several important observations from Table X. Firstly, it shows that the renewability/unlinkability criterion for most of the studied face BTP methods (\cite{b18, b19, j20, d21a, d21, e20, e22, o21, m19, d20, w20, h21, p19, p15, p16, z19, j18, jc19, j19, s21, c19, c20}) was evaluated via \textit{verbal implication}. In general, this means that renewability (or revocability/cancellability/diversity) was claimed to be possible by changing certain (external) user-specific parameters, and unlinkability was assumed due to the \textit{randomness} of these parameters; however, no theoretical or experimental proof was provided to formally validate these implications. For example, \cite{b18, b19, j20, d21a, d21, e20, e22, o21}, which used homomorphic encryption (HE) to secure face templates, implied that the renewability/unlinkability criterion can be satisfied by using different encryption keys to generate different ciphertexts (protected templates) from the same subject's face templates. Although the nature of encryption operations, in general, suggests that the resulting ciphertexts should be very different, no analysis has been provided to demonstrate the extent of these differences in the context of face template protection. Furthermore, we know that face templates protected via HE are fully invertible if the decryption key is leaked, yet this has not been considered in evaluating the renewability/unlinkability criterion.

Other examples of renewability/unlinkability evaluations via \textit{verbal implication} can be found in \cite{p15, p16, z19, j18, jc19, j19, d20}, where all the proposed face BTP methods involved training a neural network or other machine learning algorithm to map a subject's face template to a random, pre-defined binary codeword (which was then cryptographically hashed). Since the codewords were randomly generated (and further ``randomised'' by the hashing operation), they were assumed to be uncorrelated with the face templates to which they have been assigned. So, renewability/unlinkability was said to be achievable by changing the user-specific codewords and re-training the network to learn the mapping between a subject's face template and their new codeword. This appears to be a reasonable justification for fulfilment of the renewability/unlinkability criterion; however, there is no analysis to demonstrate how scalable the renewability effort would be in practice (i.e., in terms of the re-training process for each new template).  

Most of the other works that verbally implied the fulfilment of the renewability/unlinkability criterion (\cite{m19, w20, h21, p19, c19, c20}), also relied on their method's ability to update certain user-specific parameters. Unlike \cite{b18, b19, j20, d21a, d21, e20, e22, o21}, for which fulfilment of this criterion can be related to the nature of existing HE algorithms, or \cite{p15, p16, z19, j18, jc19, j19, d20}, for which adherence to the criterion is justifiable by the randomness of the pre-defined protected templates, the effect on renewability/unlinkability by changing the user-specific parameters in \cite{m19, w20, h21, p19, c19, c20} is not as readily evident. In particular, although it is usually easy to deduce that using different parameters would result in the generation of a different protected template for the same subject, it is difficult to imagine whether or not those protected templates would be unlinkable (or the extent of their unlinkability). In this case, a theoretical and/or empirical proof may be considered necessary for backing up the verbal claims. 

To complete the analysis on \textit{verbal implication} renewability/unlinkability evaluations, we should mention the approach in \cite{s21} \footnote{An ensemble network \textit{classifies} a fused face feature vector (Section II.B).}. In this work, it was claimed that the lack of template storage means that the protected template cannot be compromised and cancellability is, therefore, ensured. This cannot be considered a justification for cancellability, however, since there is insufficient analysis to prove that the trained model does not leak any information about the face templates it is used to classify. Also, diversity was said to be ensured since the trained model \textit{cannot} be used across different applications, but this is the \textit{opposite} of the accepted notion of diversity.  

The second important observation from Table X is that very few (i.e., 4) of the studied face BTP works presented a renewability/unlinkability evaluation via a \textit{theoretical proof}. Of these methods, \cite{g19} justified adherence to this criterion in terms of renewable codewords in a Fuzzy Commitment construct, which were considered as lattice points spaced sufficiently apart to avoid cross-matching of different protected templates (which consist of original templates bound with the codewords). A Fuzzy Vault scheme was adopted in \cite{r21, r22}, and a theoretical analysis was presented to prove that the probability of correlating two fuzzy vaults from the same subject using the extended Euclidean algorithm is negligible. Similarly, \cite{j21} provided an in-depth proof of the complexity of finding a link between several of a person's protected face templates, relating the justification of the method's revocability and unlinkability to its irreversibility. Finally, although the main renewability/unlinkability evaluation in \cite{l21} was actually empirical, a theoretical estimation of the number of cancellable protected templates that can be generated from a single face template was provided. This is the only one of the studied face BTP methods that provides such an estimate, although this is more likely to be an \textit{over-}estimate of the true number in practice (since it is unlikely that all possible protected templates will be sufficiently different from each other to ensure unlinkability).

The third observation from Table X is that fewer than a quarter of the face BTP works (\cite{ko20, ko21, k20, d19, x20, kh21, kh22, m21, p21, l21}) presented an \textit{empirical} evaluation of their methods' renewability/unlinkability. Most of these (\cite{k20, kh21, kh22, m21, p21, l21}) adopted the unlinkability evaluation framework from \cite{gb18}, which begins by generating several\footnote{\cite{gb18} recommends at least 5, but uses 10 in own experiments.} protected templates per subject (using different transformation parameters). Then, each protected template is compared to the \textit{same} subject's protected templates to generate a set of \textit{mated} comparison scores and to protected templates from all \textit{other} subjects to generate \textit{non-mated} scores. The relationship between the two score distributions is then used to measure the \textit{unlinkability} in terms of two metrics: $D_{\leftrightarrow}(s)$, a local score-wise measure of the degree of linkability based on the likelihood ratio between the mated and non-mated scores, and $D_{\leftrightarrow}^{\mathit{sys}}$, a global measure of the overall linkability of the protected recognition system. Both metrics produce values in the range $[0, 1]$, where 0 and 1 indicate full \textit{unlinkability} and full \textit{linkability}, respectively. An important advantage of this type of evaluation is that it defines unlinkability in a practical, readily understandable context. We must remember, however, that protected templates may be linkable in ways other than by using the comparison scores, so alternative metrics may need to be considered when evaluating the renewability/unlinkability criterion.

In addition to using the unlinkability framework from \cite{gb18}, the ``revocability'' of the face BTP method in \cite{l21} was evaluated by comparing the \textit{genuine}, \textit{impostor}, and \textit{mated-impostor} score distributions. The \textit{genuine} scores came from the normal verification scenario, where the \textit{same} user-specific transformation parameters were used for all face templates from the same subject. The \textit{impostor} and \textit{mated-impostor} scores corresponded to the \textit{non-mated} and \textit{mated} scores, respectively, in the framework from \cite{gb18}. The \textit{impostor} and \textit{mated-impostor} score distributions were found to largely overlap and be separated from the \textit{genuine} distribution, implying that protected templates from the same subject (using different transformation parameters) are as different as are protected templates from different subjects, i.e., protected templates are revocable. A similar evaluation was presented in \cite{d19, x20}.

Empirical evaluations not based on comparing score distributions, were employed in \cite{p21, ko20, ko21}. Although the unlinkability of the face BTP method in \cite{p21} was evaluated using the framework from \cite{gb18}, a separate ``cancellability'' analysis was presented based on the system's FMR when \textit{different keys} are used to generate different protected templates for the same subject. This FMR was found to be lower, in general, than the FMR computed for different identities, thereby suggesting that the protected templates are cancellable. Finally, as discussed in Section III.C, \cite{ko20, ko21} demonstrated that it is possible to invert a face template protected by the Fuzzy Commitment scheme to recover a close approximation of the corresponding face feature vector and, consequently, the face image.  The success rate of this inversion attack was used as an indicator of the BTP method's inability to satisfy the renewability/unlinkability criterion, since the face image reconstructed from one protected system could be used to access another system in which the same person is enrolled. 

The fourth and final observation from Table X is that only \cite{kh21, kh22, p21} provided a comparison of the renewability/unlinkability of their proposed method to that of other face BTP methods. In both works, this comparison was based on the global $D_{\leftrightarrow}^{\mathit{sys}}$ measure of unlinkability from \cite{gb18}, and additionally in terms of the FMR-based cancellability measure in \cite{p21}. It would be useful to see more such comparisons in the face BTP literature in the future (when deemed appropriate). 

Overall, our investigation into the renewability/unlinkability evaluation techniques among the existing face BTP methods showed that, for the most part, there is a preference for verbal statements \textit{implying} the fulfilment of this criterion in favour of an in-depth analysis \textit{proving} it. This is understandable in light of the fact that such analysis may be considered redundant when the fulfilment of the criterion seems self-evident; however, something that seems obvious in theory may not turn out to be as evident in practice (especially when certain assumptions break down). So, we would recommend a greater focus on analysis that goes the extra mile to \textit{demonstrate} fulfilment of the renewability/unlinkability criterion, especially in terms of empirical evaluations. This would help the BTP community develop a more concrete (and unified) definition of what it means to satisfy the renewability/unlinkability criterion \textit{in practice}, thereby allowing for a better understanding of how to develop BTP methods that meet this requirement.

\section{Conclusion}

This paper presented a survey of biometric template protection (BTP) methods and their evaluation techniques, when applied to face templates (images or features) in neural-network-based face recognition systems. The existing face BTP methods were categorised into \textit{Non-NN} and \textit{NN-learned} approaches, depending on what role the neural network (NN) plays in the face recognition system pipeline. Non-NN BTP methods use a neural network as a feature extractor, but not to accomplish the actual BTP part (which is based on a non-NN algorithm). On the contrary, NN-learned BTP methods use a neural network specifically for the purpose of learning a transformation from an unprotected face template to its protected version. In the literature, Non-NN face BTP methods are currently more popular than NN-learned methods, but the trend in the number of publications in each category suggests that interest in NN-learned face BTP methods is growing. The main advantage of Non-NN face BTP methods is their flexibility, since they can usually be easily incorporated into existing face recognition systems; however, the complexity of these protection mechanisms is somewhat limited by what we can algorithmically define. On the other hand, the chief attraction of NN-learned BTP methods lies in their potential to learn a complex protection algorithm without the need to explicitly define it; however, this complexity often comes at the price of less certainty about the nature of the learned protection method and thus limited understanding of how to evaluate it. So, we predicted that the face BTP field will likely evolve towards an amalgamation of these two types of approaches, to combine the advantages of a more precise BTP algorithm definition with the benefits of greater algorithm complexity. This would allow us to generate more robust protected templates, whose efficacy can be clearly evaluated.

The robustness of a face BTP method is commonly evaluated in terms of its ability to satisfy three criteria: recognition accuracy, irreversibility, and renewability/unlinkability. This survey investigated the techniques used to evaluate each of these criteria, among the existing face BTP methods. In most cases, the metrics and plots used to evaluate the recognition accuracy of protected face recognition systems were found to be the same as those used to evaluate standard (unprotected) face recognition systems. This is to be expected, since the incorporation of a BTP algorithm into a face recognition system does not change the system's aim, which is to recognise its (enrolled) users. Evaluations of the irreversibility criterion were found to be largely based on theoretical assumptions and estimations, which may result in inadequate or misleading representations of the irreversibility in practice (especially in the worst-case scenario of a fully-informed attacker, which is the most difficult threat model outlined in the ISO/IEC standard 30136 on \textit{Performance testing of biometric template protection schemes}). So, we recommended a greater focus on developing evaluation strategies that would better reflect the expected irreversibility of our BTP methods in practice. Finally, our investigation into the renewability/unlinkability evaluation techniques among the studied face BTP methods, revealed that there is an overwhelming tendency to present verbal statements \textit{implying} the fulfilment of this criterion, as opposed to providing in-depth analysis that \textit{proves} it. Although there exists an open-source unlinkability evaluation framework, thus far it has only been used by a handful of BTP researchers. So, we advocated for the investment of more effort towards developing and adopting such empirical evaluation techniques, which would help the BTP community establish a more concrete (and unified) definition of what it means to satisfy the renewability/unlinkability criterion in practice.

To complete the picture on where the face BTP field currently stands, this survey also investigated the \textit{reproducibility} of the proposed methods and their evaluations. Reproducibility was defined in terms of the public availability of: (i) the face datasets used for training and testing the proposed BTP methods, (ii) any pre-trained DNN models or known DNN architectures that were used to implement/evaluate the protected face recognition systems, and (iii) the code that ties together the entire implementation of the BTP algorithm and its evaluation. Overall, we found that, while most of the adopted datasets and DNN models/architectures are publicly available, only a few of the studied face BTP works provided a link to a public code repository containing their method implementation and evaluation procedures. So, we concluded that, at this stage, it would be difficult to faithfully reproduce most of the existing face BTP works. Although we acknowledged the possibility of private code exchanges between individual researchers, we noted that this does not benefit the BTP research community as a whole. So, we would like to finish this paper off with a call for action on making our BTP work freely and equally available to all researchers (when possible). This would help towards developing a deeper understanding of each method's characteristics, which is important for advancing the BTP field to meet the growing expectations of security and privacy preservation in practical face recognition systems.

\section*{Acknowledgment}

This work was funded by CITeR (Center for Identification Technology Research) grant 20F-01I, with affiliates IDEMIA and SICPA.

\bibliographystyle{IEEEtran}
\bibliography{bibliography}

\end{document}